\documentclass[reprint,aps,pre]{revtex4-1}
\usepackage{amsmath,amsfonts,amscd,amssymb,graphicx,amsthm}
\theoremstyle{definition}

\usepackage{subeqnar}
\usepackage{dcolumn}
\usepackage{booktabs}
\usepackage{overpic}
\usepackage{color}
\usepackage[caption=false]{subfig}
\usepackage{adjustbox}
\usepackage{rotating}
\usepackage{float}
\makeatletter
\let\newfloat\newfloat@ltx
\makeatother
\usepackage{algpseudocode}
\usepackage{placeins}
\usepackage{algorithm}
\usepackage{comment}
\usepackage{soul, xcolor}
\usepackage[hidelinks]{hyperref}
\usepackage{cleveref}
\usepackage{natbib}
\usepackage{tabularray}
\usepackage{svg}

\begin{document}

\title{Fault-tolerant shallow recurrent decoders for sparse sensing \\ with missing sensor measurements}

\author{Matteo Romilio Pasqual$^\ast$, Matteo Tomasetto$^{\ast\ast}$, J. Nathan Kutz$^\dag$, and Andrea Manzoni$^\ddag$}

\vspace{0.5cm}

\affiliation{$^\ast$ Department of Electronics, Information and Bioengineering, Politecnico di Milano, Milano, Italy}
\affiliation{$^{\ast\ast}$ Department of Mechanical Engineering, Politecnico di Milano, Milano, Italy}
\affiliation{$^{\dag}$ Autodesk Research, London, UK} 
\affiliation{$^{\ddag}$ MOX - Department of Mathematics, Politecnico di Milano, Milano, Italy}


\begin{abstract}
Real-world sensing systems are inherently imperfect: sensors may fail, intermittently disconnect, operate with asynchronous sampling rates, and be subject to noise, leading to measurement streams that are irregular, incomplete, and heterogeneous. Nevertheless, sensing strategies for real-time state reconstruction typically assume synchronous and continuously active sensors, a constraint that substantially limits their applicability in operational environments. In this work, we introduce fault-tolerant sparse sensing strategies in the context of shallow recurrent decoders (SHRED) to enable high-dimensional state estimation from sparse, asynchronous, and missing sensor measurements. Specifically, we consider naive imputation strategies, fault-tolerant recurrent temporal encoders, and we introduce four different transformer-based masking mechanisms to make SHRED explicitly aware of sensor availability through the binary data-availability mask. Moreover, we further introduce a multi-step decoding training objective for SHRED, resulting in smoother latent trajectories and more robust predictions. Throughout a synthetic benchmark and three real-world problems in hydrodynamic and ocean modeling, we show that {\em (i)} transformer-based masking mechanisms outperform standard imputation techniques and fault-tolerant recurrent architectures, {\em (ii)} the multi-step decoding training objective enhances reconstruction fidelity and robustness in the presence of heterogeneous missing-data patterns, while {\em (iii)} the computational efficiency of SHRED is preserved, with no additional inference-time overhead.
\end{abstract}
\maketitle

\section{INTRODUCTION}

Many real-world dynamical systems evolve in high-dimensional state spaces that cannot be directly observed in their entirety~\cite{ROM_2, holmes2012turbulence}. This challenge arises across a wide range of scientific and engineering applications, including fluid and solid mechanics, weather and climate modeling, plasma physics, and structural health monitoring. In these systems, sensors provide the primary interface between theory and reality, yet their deployment is inevitably limited by cost, reliability, physical inaccessibility, and operational constraints. However, being able to access local measurements is crucial to, e.g., inferring full state dynamics, designing feedback control strategies, or predicting quantities of interest.

\begin{figure*}
    \centering
    \includegraphics[width=\textwidth]{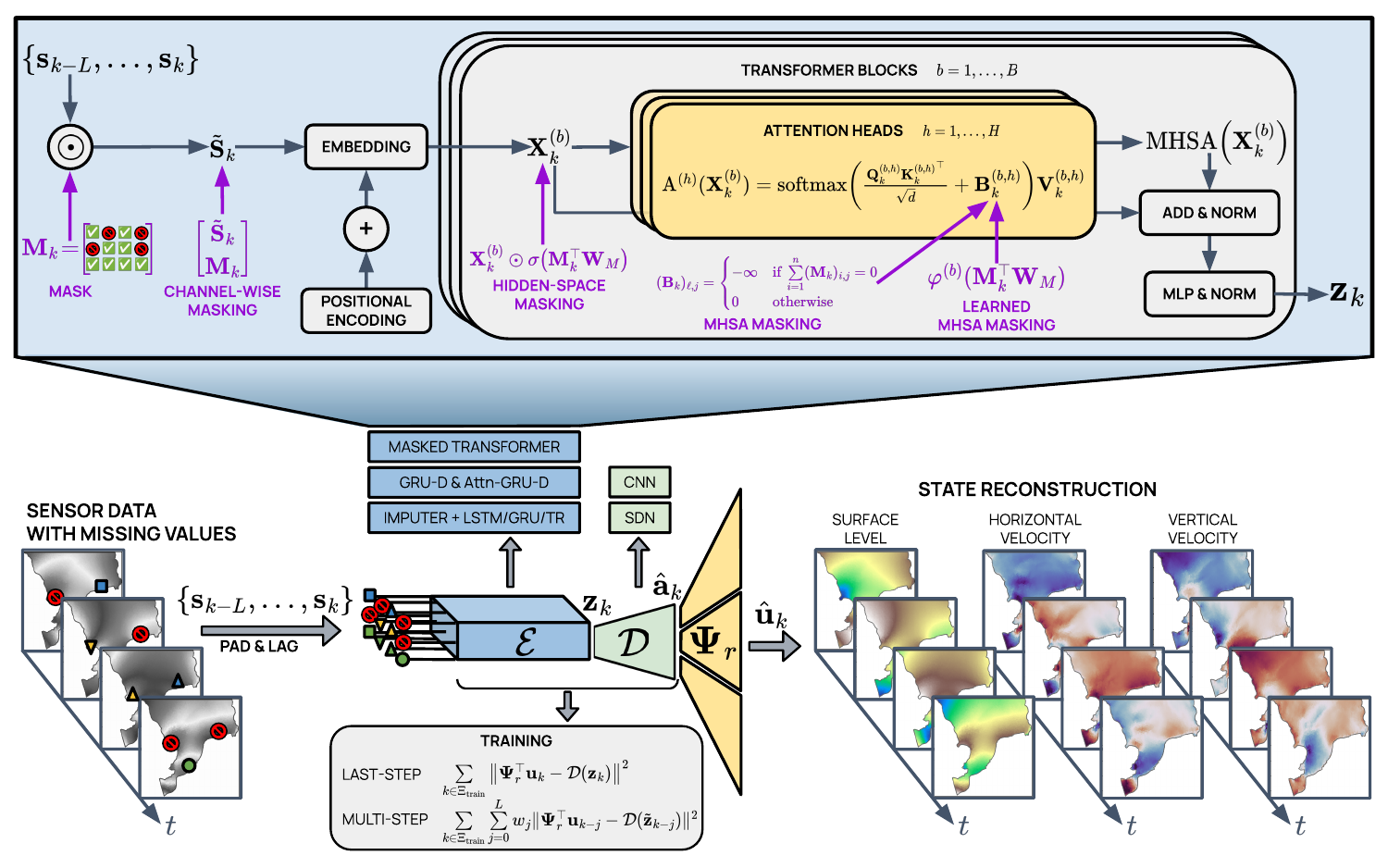}
    \caption{Overview of fault-tolerant shallow recurrent decoders for state reconstruction from sparse sensors under missing sensor measurements. Sensor measurements $\{{\mathbf{s}_{k-L},\ldots,\mathbf{s}_k}\}$ with possibly missing entries are padded, lagged, and processed by a fault-tolerant encoder $\mathcal{E}$ to produce a latent embedding $\mathbf{z}_k$, which is subsequently decoded into the POD coefficients $\hat{\mathbf{a}}_k$ via the shallow decoder $\mathcal{D}$. The state reconstruction at time $t_k$ is then given by $\hat{\mathbf{u}}_k = \boldsymbol{\Psi}_r \hat{\mathbf{a}}_k$, where $\boldsymbol{\Psi}_r$ denotes the (possibly field-specific) $r$ dominant POD modes. The upper panel illustrates the four Transformer-based masking strategies proposed where the binary data-availability mask $\mathbf{M}_k$ is injected at different levels of the encoder architecture, namely \emph{channel-wise masking}, \emph{hidden-space masking}, \emph{MHSA masking}, and \emph{learned MHSA masking}. The training box reports the standard last-step SHRED loss function and the multi-step decoding objective, where the decoder is supervised at intermediate time steps to encourage temporally consistent latent representations across the observation window.}
    \label{fig:masked-shred}
\end{figure*}

Recovering the state of a dynamical system from partial observations enables downstream tasks such as monitoring, forecasting, anomaly detection, and control, while avoiding expensive and impractical full-field measurements. Traditional approaches to state reconstruction are grounded in linear algebra frameworks. The singular value decomposition (SVD) -- also known as proper orthogonal decomposition (POD)~\cite{Sensor_P_1,kutz2013data,POD_2, Benner2015siamreview} -- provides an optimal low-rank approximation of high-dimensional state data through a linear projection onto a low-dimensional basis of spatial functions capturing most of the data variability. After computing the dominant POD modes from the training data, the coefficients of the low-rank representation can be estimated from a limited number of point measurements by solving a linear inverse problem, provided that the sensing matrix is well-conditioned~\cite{gappy_1, Willcox2006}. However, the requirement for a well-conditioned sensing matrix imposes constraints on both the number and placement of sensors, leading to an NP-hard optimization problem for identifying the optimal configuration~\cite{Cohen2003, SENSOR_P_3, SENSOR_P_5} and potentially limiting its applicability in real-world settings~\cite{SENSOR_F_3, SENSOR_F_4}. Moreover, the reconstruction accuracy of this method is bounded by the expressiveness of the POD modes used, so that fine-scales or nonlinear features of the field cannot be recovered entirely when only a few modes are retained. 

The advent of scientific machine learning has introduced new data-driven methods for addressing limitations of traditional linear methods~\cite{scientificML_1, SCIENTIFIC_ML_3, Quarteroni2025}. Neural networks and, in particular, shallow decoder networks (SDNs), leverage the universal approximation theorem of feedforward architectures~\cite{approxTH} to learn an end-to-end nonlinear mapping from sensor measurements directly to the corresponding high-dimensional state~\cite{SDN_1, SDN_2, SDN_5, luo2023}, without requiring an explicit basis representation as in POD-based methods. Similarly, the Senseiver~\cite{Santos2023} employs an attention-based architecture inspired by the Perceiver IO~\cite{perceiver} to reconstruct spatio-temporal fields at arbitrary query locations from sparse observations. Once trained on a set of snapshots, these strategies can capture nonlinear spatial correlations in the data that are inaccessible to linear methods and have been shown to outperform linear POD-based architectures in field reconstruction, particularly in ill-posed settings, i.e., with few or poorly placed sensors. Nevertheless, POD-based methods, SDNs, and the Senseiver all share an intrinsic limitation: they operate on static snapshots, treating each time instant independently and discarding the temporal dynamics of the measurement signal. The temporal history of observations, naturally available whenever sensors operate continuously over time, is essential for capturing intrinsic dynamical features, such as temporal correlations between consecutive measurements.

The shallow recurrent decoder (SHRED) framework proposed by Williams et al.~\cite{SHRED} addresses this limitation by combining a recurrent temporal encoder with a shallow spatial decoder. The encoder, which is typically modeled via recurrent neural networks such as, e.g., long short-term memory (LSTM)~\cite{LSTM, LSTM_2} or gated recurrent unit (GRU)~\cite{GRU} networks, processes a sliding window of lagged sensor measurements, compressing their temporal history into a latent representation, which is subsequently mapped to the full field by a SDN. Thanks to the temporal history of sensor readings, SHRED achieves dramatically lower reconstruction errors than classical techniques, even with a very limited number of randomly placed sensors. Different SHRED extensions have recently been proposed in the literature. For instance, Riva et al.~\cite{riva2024robuststateestimationpartial} take into account an ensemble of SHRED models to cope with noisy measurements. Moreover, the SHRED framework has been extended to parametric systems~\cite{SHRED-ROM, shred-c}, demonstrating that the temporal history of a handful of sensors suffices to reconstruct and control high-dimensional fields across varying physical and geometrical parameters. Gao et al.~\cite{gao2025sparse}, instead, introduce a regularization based on sparse identification of nonlinear dynamics (SINDy)~\cite{SINDY} to promote interpretable latent dynamics, further improving forecasting performance and explainability in challenging scenarios. More recently, Yermakov et al. employ a Transformer~\cite{attention} as temporal encoder in the so-called T-SHRED framework~\cite{TSHRED}, thus exploiting the multi-head self-attention mechanism to capture long-range temporal dependencies among lagged sensor measurements. 

SHRED has been successfully employed in a wide range of applications, while typically considering only simulated data, synchronous sensors with uninterrupted operation throughout the observation period, and pointwise field evaluations as synthetic sensor readings. The deployment of SHRED in real-world environments must account for challenging sensing conditions, including incomplete, noisy, and asynchronous data streams~\cite{TS1}. This results in time-series that are irregular, with different sampling rates, and may contain gaps and missing values~\cite{TS2}. Moreover, in practical settings, sensor measurements do not exactly match direct pointwise evaluations of the underlying state, but rather provide (possibly indirect) observations that are correlated with the system state through an observation process. Note that discarding all sensors that are not jointly available over the entire observation period severely degrades reconstruction quality, particularly in the ultra-sparse-sensor regime. Moreover, this naive strategy becomes infeasible when sensor observation windows do not overlap, as naturally occurs under fully asynchronous sampling. To overcome this limitation, CS-SHRED~\cite{SHRED_L1} and LAPIS-SHRED~\cite{LAPIS-SHERD} employ, respectively, (possibly expensive) convex minimization problems and latent-space inference to recover missing sensor values and reconstruct spatio-temporal dynamics from corrupted or incomplete observations.

In this work, we consider different efficient fault-tolerant sparse sensing strategies in the context of SHRED. First, naive imputation strategies are taken into account to infer missing sensor values, at the price of introducing additional estimation errors and potentially biasing the reconstructed field. In particular, standard baselines include mean, median, or last observation carried forward (LOCF) imputation. Rather than imputing missing sensor measurements before feeding the resulting sequence to the SHRED temporal encoder, we take into account recurrent models specifically designed to handle missing data in the context of time-series forecasting, such as GRU-D~\cite{GRUD} and its attention-augmented variant attnGRU-D~\cite{AttnLSTM}. Finally, we introduce four different transformer-based masking mechanisms to conveniently handle missing and asynchronous sensor values in the SHRED temporal encoder. Specifically, we explore both fixed and learnable masking mechanisms applied at the channel level (channel-wise masking), at the hidden space level (hidden-space masking), or at the multi-head self-attention level (fixed or learned MHSA masking). Naive missing-data imputation strategies, fault-tolerant recurrent temporal encoders and the proposed masked shallow recurrent decoders are then compared on a synthetic benchmark and on three real-world problems in hydrodynamic and ocean modeling, where sensor data are not always available or synchronous. This work thus aims at providing practical guidance for the deployment of fault-tolerant SHRED models in operational settings characterized by heterogeneous and incomplete sensor coverage.

A further contribution of this work concerns the training objective considered to fit SHRED. The temporal encoders within the original SHRED architecture and its extensions, such as T-SHRED, entail a hidden state for every time step received as input. However, only the last latent representation is passed to the decoder both during training and deployment, leaving the intermediate temporal representations entirely unexploited. This limitation is particularly consequential in the context of heterogeneous missing data: when sensor observations are irregular or partially absent, it is beneficial to leverage the temporal context accumulated across the lookback window to compensate for the missing information. Therefore, rather than relying on a SINDy regularization~\cite{gao2025sparse,TSHRED}, which requires careful dictionary selection and dedicated optimization routines, we propose a multi-step
decoding training strategy. This approach is inspired by the auxiliary classifiers in GoogLeNet~\cite{GoogleNet}, which enforce representational regularity at intermediate layers of a deep convolutional neural network by attaching additional supervision at multiple depths. Under the proposed framework, we decode intermediate hidden states of the temporal encoder using the same spatial decoder, and combine the corresponding reconstruction losses into a weighted objective, thus providing supervision across multiple temporal depths. The multi-step decoding training strategy acts as a powerful regularizer, forcing the temporal encoder to build informative latent representations throughout the entire temporal window. By ensuring that the latent trajectory remains well-structured at every step, this strategy leads to improved reconstruction accuracy and enhances the model robustness to missing data. Finally, during inference, the additional decoding heads are discarded, allowing the architecture to reduce to the standard, computationally-efficient SHRED pipeline. An overview of fault-tolerant SHRED architectures, with particular emphasis on the Transformer-based masking strategies proposed, is available in Figure~\ref{fig:masked-shred}.

This work is organized as follows. Section~\ref{sec:background} introduces the sparse sensing problem and reviews different state-of-the-art sensing methods, such as gappy POD, SDNs, and SHRED. Section~\ref{sec:masking} presents different fault-tolerant sparse sensing strategies in the context of SHRED. Specifically, it presents standard imputation techniques, fault-tolerant recurrent temporal encoders, and introduces four different Transformer-based masking mechanisms, namely channel-wise, hidden-space, MHSA, and learned MHSA masking. Moreover, it presents the multi-step decoding training objective, detailing its motivation and formulation within the SHRED training pipeline. Section~\ref{sec:results} shows the numerical results of the different fault-tolerant sensing strategies on a synthetic benchmark and three real-world problems in the context of hydrodynamic and ocean modeling. Finally, Section~\ref{sec:conclusion} summarizes the main findings, discusses the practical implications of the proposed contributions, and outlines directions for future work.
\vfill

\section{SPARSE SENSING}
\label{sec:background}

This section introduces the sensor-based state reconstruction problem, recalling different sensing methods, ranging from linear strategies to nonlinear and recurrent frameworks.

Let $u(\mathbf{x}, t)$ be the high-dimensional state of a dynamical system in the space-time domain $\Omega \times [0,T]$, where $\mathbf{x} \in \Omega \subset \mathbb{R}^D$ denotes the spatial coordinate and $t \in [0,T]$ the temporal variable, with final time $T>0$. We consider state evolutions described by means of partial differential equations (PDEs) in the form
\begin{equation}
    \dot{u} = \mathcal{N}(u, \mathbf{x}, t),
    \label{eq:pde}
\end{equation}
where $\mathcal{N}(\cdot)$ is a generic (possibly nonlinear) differential operator modeling the time derivative of the state $\dot{u}$. The governing PDE in Equation~\eqref{eq:pde} is, in general, spatially discretized through, e.g., finite elements or finite volumes~\cite{courant2008methods, kutz2013data}, yielding a finite dimensional discrete state vector $\mathbf{u}(t) \in \mathbb{R}^N$ evolving over time, where $N$ denotes the number of degrees of freedom required to capture the state dynamics. Moreover, a finite dimensional time grid $\{t_1,\ldots,t_{m}\}$ spanning $[0,T]$ is usually employed to discretize the problem over time, thus resulting in the state vectors $\mathbf{u}_k = \mathbf{u}(t_k) \in \mathbb{R}^N$ for $k=1,\ldots,m$.

In many practical scenarios, it is impossible to directly capture the full state due to a remarkably high dimension $N$, or to physical constraints. Instead, the system is often observed through $n \ll N$ sparse sensors, locally monitoring the state evolution at a few locations within the domain $\Omega$ over time $t_1,\ldots,t_m$. Formally, the state-to-sensor map is represented by the sensing operator $\mathcal{C}: \mathbb{R}^N \to \mathbb{R}^n$, mapping the full state to the corresponding sensor readings. For instance, in the simplest case of pointwise sampling, the sensing operator reduces to a matrix $\mathbf{C} \in \mathbb{R}^{n \times N}$ obtained by selecting $n$ rows of the $N \times N$ identity matrix. Note that, in general, $\mathcal{C}$ may embody a nonlinear observation process, thus allowing sensors to measure quantities that may differ from those to be reconstructed. The sensor measurements are thus given by
\begin{equation}
    \mathbf{s}_k = \mathcal{C}(\mathbf{u}_k) + \boldsymbol{\eta}_k \in \mathbb{R}^n, \quad \text{for} \; k = 1,\ldots,m,
\end{equation}
where $\boldsymbol{\eta}_k \in \mathbb{R}^n$ for $k=1,\ldots, m$ denotes measurement noise. The noise components are usually assumed to be independent and identically distributed, as they typically arise independently from each sensor measurement system rather than from correlated physical sources, and are often modeled as Gaussian.

The sparse sensing problem consists of finding a reconstruction mapping $\mathcal{R}: \mathbb{R}^n \to \mathbb{R}^N$ that accurately recovers the high-dimensional state $\mathbf{u}_k$ from the noisy, low-dimensional observations $\mathbf{s}_k$ for $k=1,\ldots,m$. In other words, the sparse sensing problem aims at finding the inverse of the sensing operator $\mathcal{C}$. However, this is generally an underdetermined and ill-posed problem, as $n \ll N$ limited and sparse observations do not uniquely determine the underlying high-dimensional state. To address this challenge, several strategies have been developed in the literature, as discussed in the following section.

\subsection{From linear to nonlinear sensing strategies}

Given the severely underdetermined nature of the state reconstruction problem from limited sensor readings, one may leverage the intrinsic low-rank structure of the data. Let us collect $m$ snapshots of the $N$-dimensional state vector at different time instances $t_1,\ldots,t_{m}$ into the so-called snapshot or data matrix
\begin{equation}
    \mathbf{X} = \begin{bmatrix} 
    | & | & \ldots & | \\
    \mathbf{u}_1 & \mathbf{u}_2 & \cdots & \mathbf{u}_{m} 
    \\  | & | & \ldots & |
    \end{bmatrix} \in \mathbb{R}^{N \times m}.
\end{equation}
The singular value decomposition (SVD), commonly referred to as proper orthogonal decomposition (POD)~\cite{POD_2, Benner2015siamreview}, provides the optimal (in a least-squares or statistical sense~\cite{EckartYoung}) data-driven low-rank approximation of the matrix $\mathbf{X}$, and thus constitutes a valuable option for addressing the reconstruction task. Specifically, the POD factorizes the snapshot matrix as
\begin{equation}
    \mathbf{X} = \mathbf{\Psi} \boldsymbol{\Sigma} \mathbf{V}^\top,
\end{equation}
where $\mathbf{\Psi} \in \mathbb{R}^{N \times N}$ is an orthogonal matrix whose columns contain the spatial modes, $\boldsymbol{\Sigma} \in \mathbb{R}^{N \times m}$ is a rectangular diagonal matrix containing the non-negative singular values in descending order, and $\mathbf{V} \in \mathbb{R}^{m \times m}$ is an orthogonal matrix whose columns contain the temporal modes. Retaining only the first $r \ll N$ columns of $\mathbf{\Psi}$ -- here denoted as $\mathbf{\Psi}_r \in \mathbb{R}^{N \times r}$ -- yields the optimal $r$-rank approximation of the state snapshots, that is
\begin{equation}
    \mathbf{u}_k \approx \mathbf{\Psi}_r \mathbf{a}_k \quad \text{for} \; k=1,\ldots,m,
\end{equation}
where $\mathbf{a}_k \in \mathbb{R}^r$ for $k=1,\ldots,m$ are the time-dependent POD coefficients. 

Instead of learning the map from sensor to state values, POD is helpful in reducing the sparse sensing problem to the estimation of the POD coefficients $\mathbf{a}_k$ from sensor readings $\mathbf{s}_k$ for $k=1,\ldots,m$, while exploiting POD to recover the original full-order state. In the case of a linear sensing operator $\mathbf{C}$, this is typically solved via the gappy POD approach~\cite{gappy_1, Willcox2006}, which considers the least-squares problems
\begin{equation}
    \hat{\mathbf{a}}_k = \arg \underset{\mathbf{a}_k}{\min} \lVert \mathbf{C}\mathbf{\Psi}_r \mathbf{a}_k - \mathbf{s}_k\rVert^2,
\label{eq:gappypod}
\end{equation}
for $k=1,\ldots,m$, where $\lVert \cdot \rVert$ denotes the Euclidean norm. The solution of Equation~\eqref{eq:gappypod} can be computed via the Moore-Penrose pseudo-inverse of the matrix $\mathbf{C}\mathbf{\Psi}_r \in \mathbb{R}^{n \times r}$, thus yielding the high-dimensional state reconstruction 
\[\hat{\mathbf{u}}_k = \mathbf{\Psi}_r \hat{\mathbf{a}}_k = \mathbf{\Psi}_r (\mathbf{C}\mathbf{\Psi}_r)^\dagger \mathbf{s}_k \quad \text{for} \; k=1,\ldots,m. \]
Despite its rigorous formulation, gappy POD is inherently constrained by the low-rank approximation induced by POD, since the reconstructed state is restricted to belong to the linear subspace spanned by the $r$ POD modes $\mathbf{\Psi}_r$, with reconstruction error lower bounded by the POD accuracy. Moreover, the quality of the reconstruction strongly depends on sensor placement, as poor sensor configurations may lead to ill-conditioned or rank-deficient sensing operators, significantly amplifying measurement noise and numerical errors~\cite{SENSOR_F_3, SENSOR_F_4}.

To overcome the limitations of linear methods, more flexible nonlinear strategies for state reconstruction have been proposed in the scientific machine learning community~\cite{scientificML_1, SCIENTIFIC_ML_3, Quarteroni2025}. Among these strategies, shallow decoder networks (SDNs) \cite{SDN_1, SDN_2, SDN_5, luo2023} replace the linear reconstruction map with a shallow, fully connected neural network, whose weights and biases are learned by minimizing the reconstruction error over the training data. Differently from linear approaches, SDNs capture nonlinear dependencies between sparse sensor measurements and the underlying state, leading to improved reconstruction performance over linear methods, particularly in highly sparse sensing regimes~\cite{SDN_2,SDN_5}. Similarly, the Senseiver~\cite{Santos2023} reconstructs spatio-temporal fields at arbitrary query locations from sparse observations, leveraging an attention-based architecture inspired by the Perceiver IO~\cite{perceiver}. However, similarly to gappy-POD, SDNs and the Senseiver operate on static snapshots, treating each time instant independently and ignoring the temporal correlation in the data, which is a core feature of sequential measurements. As a result, they fail to exploit the information contained in the history of observations, which is often crucial for accurately reconstructing dynamical systems, especially in the presence of limited and noisy data~\cite{SENSOR_P}.

\subsection{Shallow recurrent decoders}
\label{sec:background:shred}

Rather than reconstructing the full state field $\mathbf{u}_k \in \mathbb{R}^N$ directly from sensor readings $\mathbf{s}_k \in \mathbb{R}^n$ for $k=1,\ldots,m$, shallow recurrent decoders (SHRED)~\cite{SHRED} exploit a recurrent temporal encoder to learn a latent representation from a window of $L$ past sensor measurements. The architecture is therefore defined as
\begin{equation}
    \hat{\mathbf{u}}_k =
    \mathcal{D}\!\left(
        \mathcal{E}\!\left(
            \mathbf{s}_{k-L}, \dots, \mathbf{s}_k;\, \mathbf{W}_{E}
        \right);
        \mathbf{W}_{D}
    \right), \quad \text{for} \; k=1,\ldots,m
\end{equation}
where $\mathcal{E}: \mathbb{R}^{n \times (L+1)} \to \mathbb{R}^z$ is a recurrent temporal encoder parameterized by $\mathbf{W}_{E}$, which embeds the current value and the past $L$ observations of the $n$ sensors into a $z$-dimensional latent variable $\mathbf{z}_k \in \mathbb{R}^{z}$, while $\mathcal{D}: \mathbb{R}^z \to \mathbb{R}^N$ is a shallow decoder parameterized by $\mathbf{W}_{D}$, that spatially reconstructs the spatio-temporal fields of interest. Notably, pre-padding is applied -- i.e., $\mathbf{s}_k = \mathbf{0}$ for $k \leq L$ -- to avoid any burn-in period. 

To model temporal dynamics, i.e. to build the encoder $\mathcal{E}$, one may rely on recurrent neural networks such as, e.g., long short-term memory (LSTM)~\cite{LSTM, LSTM_2} or gated recurrent unit (GRU)~\cite{GRU} networks. Alternatively, T-SHRED~\cite{TSHRED} replaces the recurrent temporal encoder with a Transformer~\cite{attention}. This design choice leverages the self-attention mechanism to directly model pairwise interactions among lagged sensor measurements, enabling the network to capture long-range temporal dependencies more effectively than recurrent architectures. The spatial decoder $\mathcal{D}$, instead, maps the latent representation to the full spatial field, reconstructing $\mathbf{u}_k$ from the encoded state $\mathbf{z}_k$ for $k=1,\ldots,m$, and it is commonly modeled as a SDN with a progressively widening architecture, even though multiple alternatives are possible, such as convolutional neural networks.

The SHRED model is trained end-to-end by minimizing the reconstruction loss
\begin{equation}
\sum_{k \in \Xi_{\text{train}}}
\left\|
\mathbf{u}_k -
\mathcal{D}\!\left(
    \mathcal{E}\!\left(
        \mathbf{s}_{k-L}, \dots, \mathbf{s}_k;\, \mathbf{W}_{E}
    \right);
    \mathbf{W}_{D}
\right)
\right\|^2,
\end{equation}
where $ \Xi_{\text{train}}$ denotes the time indices in the training set. 

To enhance computational efficiency for large-scale problems, SHRED architectures usually exploit a compressive training strategy~\cite{SHRED-ROM, kutz2024shallowrecurrentdecoderreduced} where high-dimensional state snapshots are projected onto a $r \ll N$ dimensional subspace through, e.g., POD, so that the decoder estimates only the $r$ POD coefficients rather than the full state, that is
\begin{equation}
    \hat{\mathbf{u}}_k =
    \boldsymbol{\Psi}_r\,\hat{\mathbf{a}}_k =
    \boldsymbol{\Psi}_r\,\mathcal{D}\!\left(
    \mathcal{E}\!\left(
    \mathbf{s}_{k-L}, \ldots,
    \mathbf{s}_k
    \right)
    \right).
\end{equation}
Importantly, as shown in Figure~\ref{fig:masked-shred}, SHRED can be extended to multiple field reconstructions by concatenating the POD coefficient vectors associated with different physical fields into a single reduced-order representation to predict. After projecting the state snapshots, SHRED is trained by minimizing the reconstruction error at the reduced level
\begin{equation}
\sum_{k \in \Xi_{\text{train}}}
\left\|
\mathbf{\Psi}_r^\top\mathbf{u}_k -
\mathcal{D}\!\left(
    \mathcal{E}\!\left(
        \mathbf{s}_{k-L}, \dots, \mathbf{s}_k;\, \mathbf{W}_{E}
    \right);
    \mathbf{W}_{D}
\right)
\right\|^2,
\end{equation}
where $\mathbf{a}_k = \mathbf{\Psi}_r^\top\mathbf{u}_k$ for $k \in \Xi_{\text{train}}$ are the reduced state snapshots -- i.e., the POD coefficients -- in the training set. Once trained, SHRED generalizes to unseen sensor readings at new time instants, enabling reconstruction of the target field from previously unseen sequences of sensor measurements. Empirically, it has been shown that SHRED outperforms SDNs in reconstruction accuracy~\cite{SHRED}, as the recurrent temporal encoder is capable of exploiting the temporal correlations in the input sequence, leading to more informative latent representations.

\subsubsection*{Transformer shallow recurrent decoders}

T-SHRED~\cite{TSHRED} considers a Transformer~\cite{attention} as temporal encoder $\mathcal{E}$ within SHRED in order to directly model pairwise interactions among lagged sensor measurements through the self-attention mechanism. Unlike recurrent networks, which progressively aggregate past observations into a hidden representation, Transformers establish direct dependencies between all time steps within the observation window, thus enabling the model to effectively capture long-range temporal dependencies. 

Given the input sequence of lagged sensor measurements at time $t_k$
\[
\mathbf{S}_k = \begin{bmatrix}| & | & \ldots & | \\ \mathbf{s}_{k-L} & \mathbf{s}_{k-L+1} & \ldots & \mathbf{s}_k \\ | & | & \ldots & | \end{bmatrix} \in \mathbb{R}^{n \times (L + 1)},
\]
T-SHRED first considers a projection $\mathbf{S}_k^{\top}\mathbf{W}_{E} \in \mathbb{R}^{(L+1)\times z}$ of the sensor values onto a $z$-dimensional  embedding through the learnable matrix $\mathbf{W}_{E} \in \mathbb{R}^{n \times z}$, thus obtaining a sequence of $L+1$ tokens over time. Moreover, sinusoidal positional encoding~\cite{attention} is added to preserve the temporal ordering of the sequence, that is the zero-indexed matrix $\mathbf{PE} \in \mathbb{R}^{(L+1)\times z}$ having entries defined as
\[
\mathbf{PE}_{j,\ell} =
\begin{cases}
\sin\left(\dfrac{j}{10000^{\ell/z}}\right), & \text{if } \ell \text{ is even},\\
\cos\left(\dfrac{j}{10000^{(\ell-1)/z}}\right), & \text{if } \ell \text{ is odd},
\end{cases}
\]
for $j = 0,\ldots,L$ and $\ell = 0,\ldots,z-1$. The resulting matrix $\mathbf{X}_k^{(1)} = \mathbf{S}_k^{\top}\mathbf{W}_{E} + \mathbf{PE} \in \mathbb{R}^{(L+1) \times z}$ is then sequentially processed through $B$ Transformer blocks (with, typically, $B \geq 2$). A Transformer block is a fundamental building unit of the Transformer architecture that processes input representations to capture temporal dependencies between elements at different time instants in the input sequence. The core of each block is the multi-head self-attention (MHSA) mechanism, which computes $H$ attention operations -- i.e., the so-called attention heads -- in parallel, each parameterized by its own learnable weights, thus allowing each head to capture different temporal patterns and dependencies. Specifically, for the $h$-th head in the $b$-th block, the learnable projection matrices $\mathbf{W}_{Q}^{(b,h)}, \mathbf{W}_{K}^{(b,h)}, \mathbf{W}_{V}^{(b,h)} \in \mathbb{R}^{z \times d}$ are used to compute the so-called query, key, and value matrices
\begin{equation*}
\begin{aligned}
   \mathbf{Q}_k^{(b,h)} &=
    \mathbf{X}_k^{(b)}\mathbf{W}_{Q}^{(b,h)} \in \mathbb{R}^{(L+1)\times d},
    \\
    \mathbf{K}_k^{(b,h)} &=
    \mathbf{X}_k^{(b)}\mathbf{W}_{K}^{(b,h)}\in \mathbb{R}^{(L+1)\times d},
    \\
    \mathbf{V}_k^{(b,h)} &=
    \mathbf{X}_k^{(b)}\mathbf{W}_{V}^{(b,h)}\in \mathbb{R}^{(L+1)\times d}
    \end{aligned}
\end{equation*}
from the block input $\mathbf{X}_k^{(b)} \in \mathbb{R}^{(L+1) \times z}$. The attention operation is then defined as
\[ 
\mathrm{A}^{(h)}\!\left(\mathbf{X}_k^{(b)}\right)
    = \mathrm{softmax}\!\left(
    \frac{
    \mathbf{Q}_k^{(b,h)}
    {\mathbf{K}_k^{(b,h)}}^{\top}
    }{\sqrt{d}}
    \right)
    \mathbf{V}_k^{(b,h)},
\]
where the softmax function is applied row-wise. Specifically, the product $\mathbf{Q}_k^{(b,h)}{\mathbf{K}_k^{(b,h)}}^{\top} \in \mathbb{R}^{(L+1)\times(L+1)}$ collects the pairwise dot-product scores between all query and key vectors, which quantify how strongly each pair of time instants is aligned. The softmax function converts these scores into positive attention weights, with each row summing to one. Finally, the attention weights are multiplied by the value matrix $\mathbf{V}_k^{(b,h)}$, so that the output at each time instant is a weighted combination of the value vectors, with input-dependent weights determined by the learned query and key projections. The outputs of the $H$ attention heads are concatenated and linearly projected through a learnable matrix $\mathbf{W}_{O} \in \mathbb{R}^{H d \times z}$, yielding the MHSA representation
\begin{equation*}
    \mathrm{MHSA}\!\left(\mathbf{X}_k^{(b)}\right)
    =
    \begin{bmatrix}
    
    \mathrm{A}^{(1)}\!\left(\mathbf{X}_k^{(b)}\right),
    \ldots,
    \mathrm{A}^{(H)}\!\left(\mathbf{X}_k^{(b)}\right)  \end{bmatrix}
   \mathbf{W}_{O}.
\end{equation*}
Finally, the Transformer block applies residual connections and layer normalization~\cite{ba2016layernormalization} -- i.e., each input feature is standardized to zero mean and unit variance, and learnable per-feature scale and shift parameters are applied to stabilize training -- around both the MHSA module and the (position-wise) multi-layer perceptron (MLP), that is
\begin{equation*}
\begin{aligned}
    \tilde{\mathbf{X}}_k^{(b)} &=
    \mathrm{LayerNorm}\!\left(
    \mathbf{X}_k^{(b)}
    +
    \mathrm{MHSA}\!\left(\mathbf{X}_k^{(b)}\right)
    \right),
    \\
    \mathbf{X}_k^{(b+1)} &=
    \mathrm{LayerNorm}\!\left(
    \tilde{\mathbf{X}}_k^{(b)}
    +
    \mathrm{MLP}\!\left(\tilde{\mathbf{X}}_k^{(b)}\right)
    \right).
\end{aligned}
\end{equation*}
The output of the last Transformer block $\mathbf{X}_k^{(B+1)} \in \mathbb{R}^{(L+1) \times z}$ can be regarded as a collection of $L+1$ latent representations over the observation window. To recover the full state field at time $t_k$, its last row $\mathbf{z}_k =\left(\mathbf{X}_k^{(B+1)}\right)^\top  \mathbf{e}_{L+1} \in \mathbb{R}^z$, where $\mathbf{e}_{L+1} \in \mathbb{R}^{L+1}$ is the $(L+1)$-th one-hot selection vector, has to be upscaled by the spatial decoder $\mathcal{D}$, which is typically modeled as a SDN or a convolutional neural network.


\section{FAULT-TOLERANT SHALLOW RECURRENT DECODERS}
\label{sec:masking}

In this section, we introduce different fault-tolerant sparse sensing strategies in the context of SHRED. In particular, we consider naive imputation techniques to recover missing sensor values, fault-tolerant recurrent temporal encoders, and we present four different Transformer-based masking mechanisms that enable SHRED to deal with irregular and asynchronous time-series data, such as sensor readings with missing values and different sampling rates. 

\subsection{Handling missing sensor measurements}
\label{sec:background:missingvalues}

\begin{figure*}
    \centering
    \includegraphics[width=\linewidth]{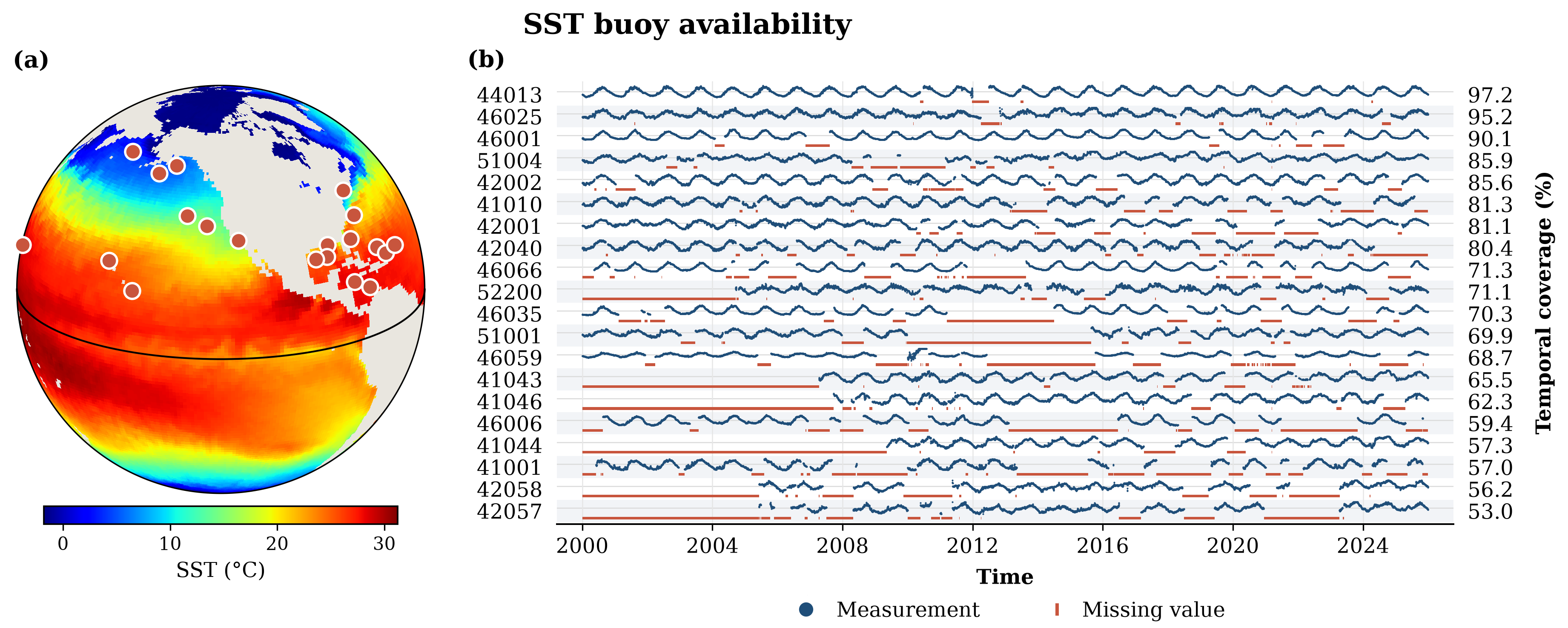}

    \caption{{\em Sea surface temperature}. {\bf (a)} Spatial coverage of moored buoys collecting in situ sea surface temperature (SST) measurements~\cite{sst_buoys}. {\bf (b)} Sensor data availability over time of different moored buoys~\cite{sst_buoys}, covering the period from January 2000 to December 2025, showing availability rates ranging from $97.2\%$ to $53.0\%$ and different data availability patterns. The available sensor readings are depicted by blue dots, while missing values are represented by red bars.}
    \label{fig:availability}
\end{figure*}
The SHRED framework described in Section~\ref{sec:background:shred} assumes that all the $n$ input sensor measurements are available at every time step within the observation window $\{t_{k-L}, \ldots, t_k\}$. Under this assumption, the encoder receives a fully populated input sequence at each reconstruction step. This hypothesis is reasonable when testing the model in synthetic settings, although it is not fully attainable in real-world systems. In practice, sensors are subject to a broad spectrum of failure modes: power outages, communication dropouts, hardware degradation, scheduled maintenance, calibration windows, and environmental interference. These events produce time-series that are irregular, asynchronous across sensors, and punctuated by gaps at isolated time instants or contiguous intervals with variable duration. Moreover, when multiple heterogeneous sensors are combined into a single network, their temporal availability patterns are rarely aligned, resulting in a measurement matrix with structured, sensor-dependent patterns of missing entries that reflect the operational history of each individual instrument. For instance, Figure~\ref{fig:availability} shows the spatial coverage and the sensor data availability of $20$ moored buoys collecting in situ sea surface temperature (SST) measurements~\cite{sst_buoys}, with availability rates ranging from $97.2\%$ to $53.0\%$ and different data availability patterns. In this work, we aim at considering fault-tolerant sparse sensing strategies capable of dealing with missing and asynchronous sensor values, such as defective sensors with heterogeneous sampling frequencies. 

To address incomplete, irregular and asynchronous sensor data, one may consider naive imputation strategies, where the mean, the median or the last-value of the available training data are employed to fill missing entries, with no modification to the SHRED architecture. Specifically, whenever the $i$-th sensor presents a missing value at time $t_k$, the mean imputation replaces the $i$-th entry $s_{i,k}$ in the vector $\mathbf{s}_k$ with the average of the available sensor observations over the training set, that is
\[
 \hat{s}_{i,k} = \frac{1}{|\Xi_i|} \sum_{j \in \Xi_i} s_{i,j},
\]
where $\Xi_i$ is the subset of training time steps at which sensor $i$ is observed, with cardinality $|\Xi_i|$. Similarly, the median imputation considers the median of the available observations in the training set, that is
\[
    \hat{s}_{i,k} = \mathrm{median}\{s_{i,j} : j \in \Xi_i\}.
    \label{eq:median_kmputation}
\]
Finally, the last observation carried forward (LOCF) imputation replaces missing entries with the most recent available measurement of the same sensor
\[
    \hat{s}_{i,k} = s_{i,j^*}, \; \; j^* = \max\{j < k : j \in \Xi_i \},
\]
while mean imputation is employed whenever no past sensor values are available. Thanks to local temporal continuity, LOCF is particularly effective for short, isolated gaps in smooth signals, while its performance degrades for long gaps or rapidly varying signals. All three imputation strategies can be readily combined with any temporal encoder architecture, since they operate entirely as pre-processing steps. 

In addition to naive imputation strategies, it is possible to consider SHRED temporal encoders $\mathcal{E}$ specifically designed to handle missing data in the context of multivariate time-series forecasting, such as GRU-D~\cite{GRUD} and its attention-augmented variant attnGRU-D~\cite{AttnLSTM}. GRU-D considers learnable temporal decay mechanisms that account for the time elapsed since the last observation of each sensor. In particular, missing input values are imputed by decaying the last observed value toward a feature-specific empirical mean, while the hidden state is simultaneously decayed toward a default state. These decay mechanisms allow the model to learn how rapidly sensor information becomes outdated and enable the elapsed time since the last observation to directly influence the recurrent dynamics. The resulting input and hidden state are then processed by a standard GRU, allowing GRU-D to jointly exploit the observed values, their missingness patterns, and the temporal structure of the data. Instead, attnGRU-D extends GRU-D by incorporating an attention mechanism to selectively weight the recurrent hidden representations across time, thus allowing the model to identify and emphasize the time steps that are most relevant for the prediction task.

\subsection{Transformer-based masking strategies}
\label{sec:masking:mshred}

Rather than relying on naive imputation techniques and fault-tolerant recurrent encoders, which address data missingness through pre-processing operations or via modified recurrence equations, in this section we incorporate data availability information directly into the Transformer encoder of T-SHRED. Specifically, we inject the binary data-availability mask at different levels of the architecture, in order to condition the internal representation of the temporal encoder both on the observed values and on their corresponding availability patterns. Doing so, the model can learn to discount missing entries instead of treating placeholder values as genuine data, which would otherwise bias the reconstruction.

Let the time grid $\{t_1, \ldots, t_m\}$ be the collection of the sampling instants of all the available sensors. In the case of missing and asynchronous sensor measurements, the observed data at reconstruction step $k$ read as follows
\begin{equation}
    \tilde{\mathbf{S}}_k = \mathbf{S}_k \odot \mathbf{M}_k,
\end{equation}
where $\odot$ denotes element-wise multiplication, $\mathbf{S}_k = [\mathbf{s}_{k-L}, \ldots, \mathbf{s}_k] \in \mathbb{R}^{n \times (L+1)}$ is the matrix collecting the sequence of sensor values over the past $L$ steps until time $t_k$ for $k=1,\ldots,m$, with possibly missing entries, while $\mathbf{M}_k\in \{0,1\}^{n \times (L+1)}$ is the mask matrix encoding the availability of each sensor over the lag window. Specifically, the mask is a binary matrix where each entry indicates the data availability of sensor $i$ at time $t_j$, that is
\[
(\mathbf{M}_k)_{i,j} = \begin{cases}
1 & \text{if sensor } i \text{ at time } t_{k+j-(L+1)} \text{ is available} \\
0 & \text{otherwise,}
\end{cases}
\]
for $i=1,\ldots,n$ and $j=1,\ldots,L+1$. The pattern of missing entries jointly depends on the sampling rates and the failure processes of the individual sensors, making the mask $\mathbf{M}_k$ generally heterogeneous across sensors and non-stationary over time. 

Differently from the original SHRED implementation tailored to regular and complete data streams, in this section we propose masked shallow recurrent decoders where we effectively inform SHRED on missing sensor values through Transformer-based masking strategies. Specifically, we take into account the observed data matrix $\tilde{\mathbf{S}}_k$ as input for $k=1,\ldots,m$, and we inject the corresponding binary mask $\mathbf{M}_k$ at different levels of the architecture. In the following, we introduce four different masking strategies: {\em channel-wise masking}, where the mask is concatenated to the input prior to embedding; {\em hidden-space masking}, which gates intermediate hidden representations via a learned projection of the mask; {\em MHSA masking}, which enforces hard attention suppression of time steps with no measurements available; and {\em learned MHSA masking}, which learns head-specific additive attention biases conditioned on the missingness pattern. The resulting Transformer-based temporal encoders $\mathcal{E} : \mathbb{R}^{n \times (L+1)} \times \{0,1\}^{n \times (L+1)} \rightarrow \mathbb{R}^z$ are thus able to return a latent representation $\mathbf{z}_k$ for $k=1,\ldots,m$ aware of the sensor availability, and allow SHRED to generalize to irregular and asynchronous data streams. 

\vspace{0.25cm}
\paragraph*{Channel-wise masking}

The first strategy to make the Transformer encoder aware of missing values is to concatenate the binary mask $\mathbf{M}_k$ directly to the observed data $\tilde{\mathbf{S}}_k$ as an additional set of feature channels prior to the input embedding layer, that is
\[
\begin{bmatrix}
  \tilde{\mathbf{S}}_k \\  \mathbf{M}_k
\end{bmatrix} \in \mathbb{R}^{2n \times (L+1)}.
\]
This effectively doubles the number of input channels from $n$ to $2n$, where the first $n$ rows correspond to the observed sensor values, while the remaining $n$ rows encode the corresponding binary availability indicators. The augmented input is then processed as in the standard T-SHRED framework. Notably, no modification is made to the attention mechanism or any other architectural component, as the mask information is injected exclusively at the input level. As a result, all subsequent layers must implicitly learn how to interpret the additional input information and how to propagate it through the encoder. The channel-wise strategy is therefore simple and architecture-agnostic, but it does not provide an explicit mechanism to prevent missing entries from influencing the attention scores, so that the model must learn to discount them implicitly during training.

\vspace{0.25cm}
\paragraph*{Hidden-space masking}
Instead of providing the binary data-availability mask as additional input, the second masking strategy operates at the level of the hidden representation. We recall that the masked sensor measurements $\tilde{\mathbf{S}}_k$ are embedded through the projection matrix $\mathbf{W}_{\text{E}} \in \mathbb{R}^{n \times z}$, while positional encoding is added, thus yielding the matrix $\mathbf{X}_k^{(1)} \in \mathbb{R}^{(L+1)\times z}$. At the beginning of each Transformer block, the hidden representation $\mathbf{X}_k^{(b)} \in \mathbb{R}^{(L+1) \times z}$ for $b = 1,\ldots,B$ is modulated element-wise by a continuous gating signal, that is
\begin{equation*}
    \mathbf{X}_k^{(b)} \leftarrow \mathbf{X}_k^{(b)} \odot \sigma\!\left(\mathbf{M}_k^\top \mathbf{W}_{M}\right),
\end{equation*}
where $\mathbf{W}_{M} \in \mathbb{R}^{n \times z}$ is a learnable projection matrix shared across different blocks, while $\sigma(\cdot)$ is the sigmoid function. The gated input is then fed to the Transformer block.

This masking mechanism explicitly amplifies or attenuates each hidden dimension at each time step according to which sensors are available, so that the model can learn to down-weight the hidden features that mainly depend on missing measurements. Moreover, since the mask directly influences the intermediate representations at every Transformer block, it ensures that the data availability information is propagated throughout the depth of the network rather than being confined to the input embedding stage, thus effectively controlling the information flow as it propagates through the model. Note that, since $\mathbf{W}_M$ is learned jointly with all the other model parameters, the resulting gating signal is not a fixed binary mask, but a continuous, data-dependent weighting. This allows the model to adaptively determine how much each hidden dimension should be suppressed or preserved at every time step, depending on the observed missingness pattern and its interaction with the hidden state.

\vspace{0.25cm}
\paragraph*{MHSA Masking}

The third masking strategy operates directly within the scaled dot-product attention mechanism, analogously to the masking mechanism considered in the standard Transformer~\cite{attention}, preventing missing values from contributing to the attention computation. Rather than modifying the input or the hidden representations, this approach introduces an additive bias to the attention scores that explicitly excludes time steps with no sensor values available from the softmax computation. Specifically, we consider the modified self-attention operation 
\begin{equation*}
    \mathrm{A}^{(h)}(\mathbf{X}_k^{(b)}) =
    \mathrm{softmax}\!\left(
    \frac{\mathbf{Q}_k^{(b,h)}{\mathbf{K}_k^{(b,h)}}^\top}{\sqrt{d}} + \mathbf{B}_k
    \right)\mathbf{V}_k^{(b,h)},
\end{equation*}
where the attention bias matrix at the $k$-th reconstruction step $\mathbf{B}_k \in \mathbb{R}^{(L+1) \times (L+1)}$ is defined as
\begin{equation}
    (\mathbf{B}_k)_{l,j} =
    \begin{cases}
        -\infty & \text{if } \sum_{i=1}^{n} (\mathbf{M}_k)_{i,j} = 0 \\
        0 & \text{otherwise},
    \end{cases}
\end{equation}
for $l,j = 1,\ldots,L+1$.  Note that the attention bias matrix is shared across Transformer blocks and attention heads. After softmax normalization, this ensures that the attention weights assigned to time steps with no sensor values available (i.e., $\sum_{i=1}^{n} (\mathbf{M}_k)_{i,j} = 0$) are exactly zero, independently of the learned projections. Note that one may additionally require that $(\mathbf{B}_k)_{l,j} = -\infty$ for all $l < j$, thereby recovering the autoregressive structure of standard Transformer models~\cite{attention}.

Importantly, since the mask is derived directly from the binary availability pattern and applied uniformly across all attention heads and Transformer blocks, this strategy imposes a hard structural constraint on information flow within the model. Note that, while time steps with no sensor values available are completely excluded from the attention computation, time instants with partial missingness are not explicitly modeled in this formulation. The MHSA masking thus yields a strict and interpretable mechanism for filtering missing information, but it does not capture finer-grained patterns of partial observability within a timestep.

\vspace{0.25cm}
\paragraph*{Learned MHSA Masking}

The fourth masking strategy extends the MHSA masking by replacing the fixed, rule-based mechanism for filtering missing information with a learnable data-dependent attention bias. Specifically, we consider the attention operation
\begin{equation*}
\text{A}^{(h)}(\mathbf{X}_k^{(b)}) = \text{softmax}\!\left(
\frac{\mathbf{Q}_k^{(b,h)}{\mathbf{K}_k^{(b,h)}}^\top}{\sqrt{d}} 
+ {\mathbf{B}_k}^{(b,h)}
\right)\mathbf{V}_k^{(b,h)},
\end{equation*}
where the attention bias ${\mathbf{B}_k}^{(b,h)} \in \mathbb{R}^{(L+1)\times(L+1)}$ for the $b$-th Transformer block and the $h$-th attention head is given by a lightweight two-layer MLP with ReLU activation function. In particular, the block-specific feedforward neural network $\varphi^{(b)}:\mathbb{R}^{(L+1) \times z} \to \mathbb{R}^{(L+1) \times (L+1) \times H}$ returns, after a suitable reshape, the attention biases ${\mathbf{B}_k}^{(b,h)} \in \mathbb{R}^{(L+1)\times(L+1)}$ for $h=1,\ldots,H$, that is
\begin{equation*}
\begin{bmatrix}
{\mathbf{B}_k}^{(b,1)}, \ldots, {\mathbf{B}_k}^{(b,H)}    
\end{bmatrix} = \varphi^{(b)}(\mathbf{M}_k^\top \mathbf{W}_M),
\end{equation*}
where $\mathbf{W}_M \in \mathbb{R}^{n \times z}$ is a learnable embedding matrix, shared across different Transformer blocks.

In contrast to the previous masking strategies, the learned MHSA masking allows each Transformer block and attention head to adjust the attention between pairs of time instants according to the data-availability pattern. This gives the model the flexibility to handle partially observed time steps, and lets different heads specialize in different missingness patterns, at the price of a soft rather than guaranteed exclusion of missing entries from the attention computation. It is important to note that the different masking mechanisms proposed are not mutually exclusive, and they can be readily combined in hybrid configurations, thus allowing the model to jointly exploit explicit availability information and more flexible, layer-wise adaptations to missing-data patterns.

\subsection{Multi-step decoding training strategy}
\label{sec:masking:lagged}

This section introduces the multi-step decoding training strategy designed to enhance latent trajectories and SHRED predictions. The original SHRED implementation, as well as its extensions such as T-SHRED, embeds a sequence of lagged sensor measurements $\mathbf{s}_{k-L},\ldots,\mathbf{s}_k$ into a single latent representation $\mathbf{z}_k \in \mathbb{R}^z$, which is spatially decoded into the full state variable. However, temporal encoders internally produce a sequence of $z$-dimensional latent representations $\tilde{\mathbf{z}}_{k-L}, \ldots, \tilde{\mathbf{z}}_{k-1},  \mathbf{z}_k \in \mathbb{R}^z$ over time. Note that, while recurrent temporal encoders such as, e.g., LSTM and GRU recursively update the hidden state over
time, Transformer-based encoders aggregate the latent representations in the rows of the output matrix $\mathbf{X}^{(B+1)}_k \in \mathbb{R}^{(L+1) \times z}$. Discarding all the hidden states but the last one may be a limiting factor when dealing with heterogeneous missing data, as temporal context accumulated across the lookback window may compensate for the missing information. Instead, supervising the decoding of intermediate latent states provides a training signal at every time step of the lookback window, which encourages the encoder to retain state information over time and makes the latent trajectory less sensitive to missing measurements at individual steps, which may be helpful in improving reconstruction accuracy and forecasting capabilities. 

To mitigate this limitation, we introduce a multi-step decoding training strategy, inspired by the auxiliary classifier mechanism introduced in GoogLeNet~\cite{GoogleNet}, where auxiliary branches are attached at intermediate depths of the network and produce additional classification predictions during training. Specifically, while training SHRED, we apply the same spatial decoder $\mathcal{D}$ to the latent variables $\tilde{\mathbf{z}}_{k-L},\ldots,\tilde{\mathbf{z}}_{k-1}, \tilde{\mathbf{z}}_k = \mathbf{z}_k$, thus obtaining multiple lagged state reconstructions. The encoder and decoder parameters $\mathbf{W}_E$ and $\mathbf{W}_D$ can then be determined by minimizing the training loss
\begin{equation*}
\sum_{k \in \Xi_{\text{train}}}
\sum_{j=0}^{L}
w_j
\lVert
\mathbf{u}_{k-j} - \mathcal{D}(\tilde{\mathbf{z}}_{k-j})
\rVert^2,
\end{equation*}
where the coefficients $w_j \in (0,1)$ for $j = 0,\ldots,L$ balance the contribution of each prediction horizon. In the following, we consider weights in the form $w_j = \alpha^{j}$ for $j = 0,\ldots,\tilde{L}$ and $w_j = 0$ for $j=\tilde{L}+1,\ldots,L$, with $0<\tilde{L}<L$ and $\alpha \in (0,1)$, so as to prioritize accuracy on more recent time instants.

Whenever a compressive training strategy is employed through, e.g., a POD-based projection of the full state fields in the training set, the multi-step decoding loss reads as follows 
\begin{equation*}
\sum_{k \in \Xi_{\text{train}}}
\sum_{j=0}^{L}
w_j
\lVert
\mathbf{\Psi}_r^\top\mathbf{u}_{k-j} - \mathcal{D}(\tilde{\mathbf{z}}_{k-j})
\rVert^2,
\end{equation*}
where the spatial decoder aims at reconstructing the lagged POD coefficients starting from the corresponding latent representations, rather than predicting the full state.

It is worth noting that, after training, only the last latent representation $\mathbf{z}_k$ is decoded to obtain the state reconstruction $\hat{\mathbf{u}}_{k}$, thus recovering the standard SHRED pipeline at inference time with no additional computational overhead.

\section{NUMERICAL RESULTS}
\label{sec:results}

This section presents the experimental validation of the proposed fault-tolerant shallow recurrent decoder strategies for handling heterogeneous missing sensor data in four different scenarios. In particular, we consider a first synthetic benchmark dealing with shallow water equations solution data, in which sensor malfunctions are artificially introduced. Moreover, we take into account three real-world sparse sensing problems in the context of hydrodynamic and ocean modeling with missing and asynchronous sensor measurements. 

Throughout the test cases, we split the temporal snapshots into training, validation and test set, so that we test the performance of fault-tolerant sensing strategies on sensor measurements unseen during training. The test set includes both data within the temporal range covered by the training data, to assess interpolation performance, and beyond the training period, to evaluate extrapolation to future time instants. Moreover, we quantitatively assess the quality of the reconstruction on test snapshots through two standard metrics: the root mean squared error (RMSE)
\begin{equation*}
    \sqrt{\frac{1}{|\Xi_{\text{test}}|} \sum_{k \in \Xi_{\text{test}}} 
    \lVert\mathbf{u}_k - \hat{\mathbf{u}}_k\rVert^2},
\end{equation*}
and the range-normalized mean absolute error (NMAE)
\begin{equation*}
    \frac{1}{|\Xi_{\text{test}}|} \sum_{k \in \Xi_{\text{test}}} \dfrac{
    \lVert\mathbf{u}_k - \hat{\mathbf{u}}_k\rVert_1}{\mathbf{u}_{\max}-\mathbf{u}_{\min}},
\end{equation*}
where $\lVert \cdot \rVert_1$ denotes the $L^1$ (Manhattan) norm, while $\mathbf{u}_{\max}$ and $\mathbf{u}_{\min}$ are the maximum and minimum state components over the test set, respectively. Note that, when reconstructing multiple physical fields, the error metrics are computed separately for each field and then averaged to obtain the final score. Finally, we consider the same training configuration for all the models and test cases, which takes into account $200$ epochs, batch size equal to $32$, AdamW optimizer with cosine annealing learning rate scheduler, and multi-step decoding training strategy with $\alpha = 0.8$ and $\tilde{L} = 16$.

\subsection{Shallow water equations}
\label{sec:results:swe}

\begin{figure*}
    \centering
    \includegraphics[width=\linewidth]{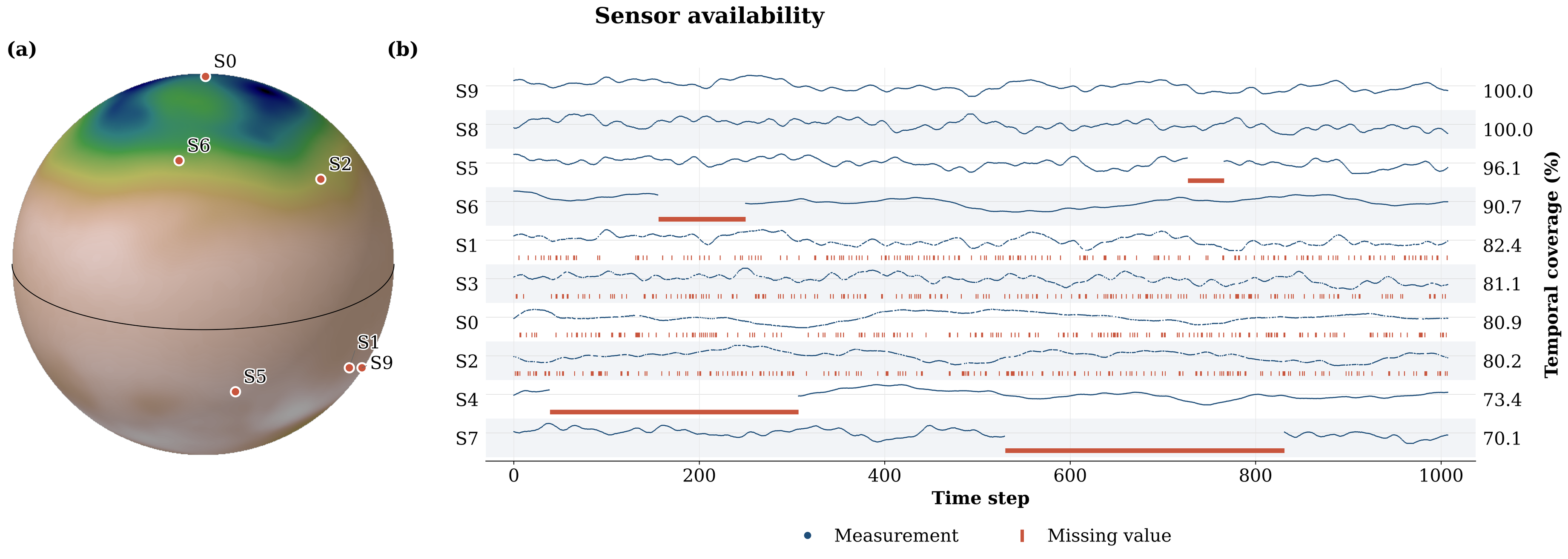}
    \caption{{\em Shallow water equations}. {\bf (a)} Spatial coverage of simulated water level stations. {\bf (b)} Sensor data availability over one year of different water level stations, showing availability rates ranging from $100\%$ to $70.1\%$ and different data availability patterns. The available sensor readings are depicted by blue dots, while missing values are represented by red bars.}
    \label{fig:swe_availability}
\end{figure*}
\begin{figure*}
    \centering
    \includegraphics[width=\linewidth]{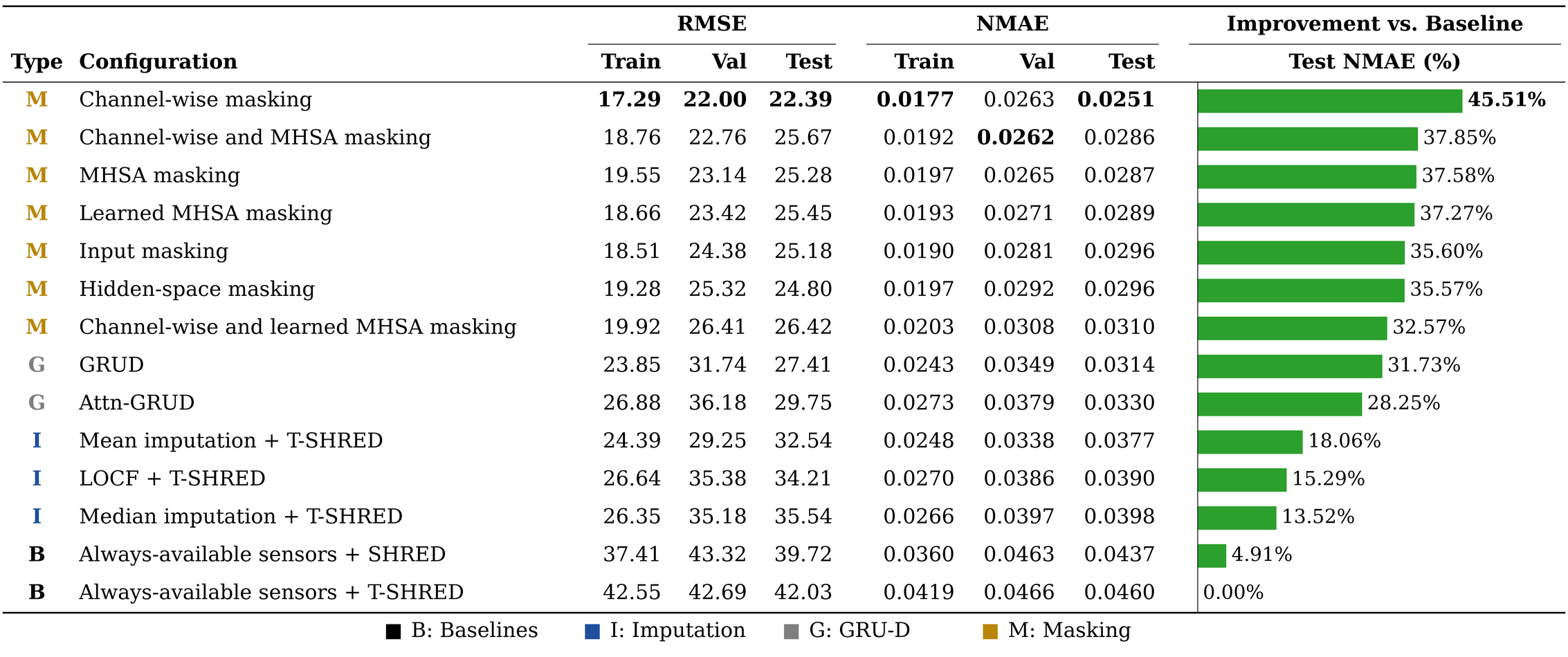}
    \caption{{\em Shallow water equations}. Root mean squared errors and normalized mean absolute errors on training, validation and test data achieved by the different competing models.}
    \label{fig:swe_table}
\end{figure*}
\begin{figure*}
    \centering
    \includegraphics[width=\linewidth]{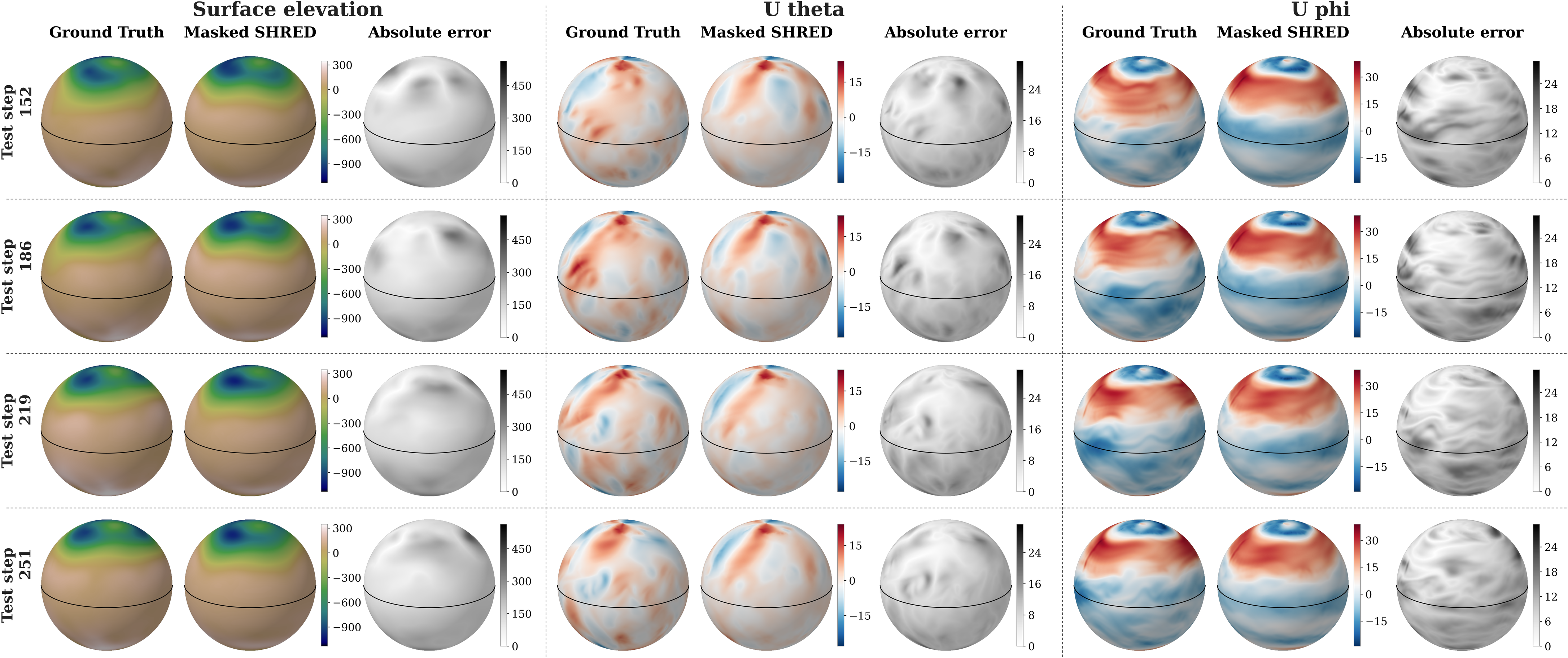}
    \caption{{\em Shallow water equations}. Ground truth, reconstruction provided by masked SHRED with channel-wise masking and reconstruction error on surface elevation (first column), polar velocity (second column) and azimuthal velocity (third column) at four different time instants in the test set.}
    \label{fig:swe_results}
\end{figure*}

The first test case serves as a numerical benchmark to test the different fault-tolerant strategies proposed in Section~\ref{sec:masking}. Specifically, we focus on high-dimensional temporal data provided by {\em The Well} dataset~\cite{ohana2024thewell} corresponding to the solution of shallow water equations on a sphere with earth-like topography and daily/annual periodic external forces. The dataset collects $1008$ observations over a year of three high-dimensional fields ($N=131072$), namely the height deviation of pressure surface from its mean height, and the velocity components with respect to the polar and azimuthal angles. We consider $10$ in situ measurement stations placed randomly in the domain, where the sensor readings are directly extracted from the numerical simulation as pointwise evaluations. To simulate different sensor behaviors, we consider $2$ always-available sensors, $4$ sensors with missing values at random time instants, and $4$ sensors with gaps over contiguous intervals with variable duration, as visible in Figure~\ref{fig:swe_availability}.

We split the data into training, validation and test set with ratio $80:10:10$, so that the test set includes both months within the temporal range covered by the training data and months beyond the training period. We then employ POD to reduce the dimensions of the fields to reconstruct: $100$ POD modes are enough to compress each field, namely the surface elevation and the velocity components, with a residual energy below $0.1 \%$. Moreover, we consider a lag equal to $100$ time steps to capture the history of the system and the temporal pattern of the sensor readings. 

To solve the reconstruction problem, we consider the different fault-tolerant sensing strategies introduced in Section~\ref{sec:masking}, such as imputation-based and GRU-D-based strategies, as well as different masking strategies in the context of T-SHRED. Moreover, we also take into account SHRED with LSTM-based recurrent temporal encoder and T-SHRED trained exclusively on the $2$ always-available sensor stations. To fairly evaluate the performance of the competing models, we consider the same training setup and architectures for all the competing models. Specifically, we employ recurrent temporal encoders with a single hidden layer having $32$ neurons, Transformer-based encoders with $1$ block, $4$ attention heads, embedding dimension equal to $32$, and a MLP with a single hidden layer of size $48$. The spatial decoder is instead modeled via a SDN with a single hidden layer having $64$ neurons, while dropout with rate $0.1$ is applied to prevent overfitting.

Figure~\ref{fig:swe_table} quantitatively assesses the performance of the different baselines and fault-tolerant sensing strategies considered through the RMSE and the NMAE on training, validation and test data, with Transformer-based masking strategies providing the most accurate results. In particular, the best-performing configuration employs channel-wise masking to obtain a NMAE on test data equal to $2.51 \%$, with a relative improvement compared to the T-SHRED baseline fitted on always-available sensors only equal to $45.51 \%$. Imputation- and GRU-D-based methods, instead, yield higher errors, although they still outperform the baselines. Notably, hyperparameter tuning and repeated training runs with different random initializations may lead to slight changes in the ranking of the top-performing models. Nevertheless, masked SHRED strategies systematically outperform imputation- and GRU-D-based methods, although no single masking configuration consistently dominates across all runs. Figure~\ref{fig:swe_results}, instead, qualitatively compares the reconstructed fields provided by the best-performing model with the ground truth at four different time instants in the test set, showing the accuracy of the proposed strategy to predict the spatio-temporal patterns of the three high-dimensional fields of interest, despite the limited and irregular input information.
\vfill 

\subsection{Southern North Sea hydrodynamics}
\label{sec:results:hydro}

\begin{figure*}
    \centering
    \includegraphics[width=\linewidth]{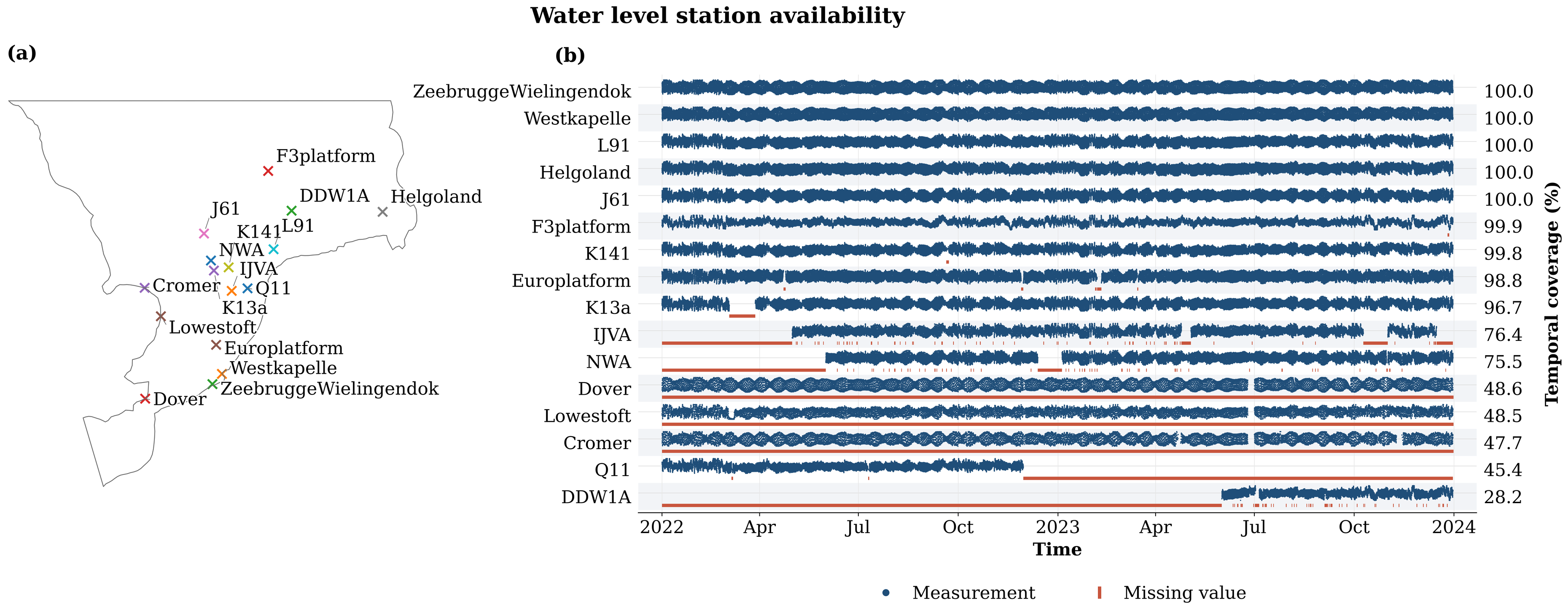}
    \caption{{\em Southern North Sea hydrodynamics}. {\bf (a)} Spatial coverage of in situ water level stations in the Southern North Sea. {\bf (b)} Sensor data availability over time of different in situ water level stations in the Southern North Sea, covering the period from January 2022 to October 2023, showing availability rates ranging from $100\%$ to $28.2\%$ and different data availability patterns. The available sensor readings are depicted by blue dots, while missing values are represented by red bars.}
    \label{fig:hydro_availability}
\end{figure*}
\begin{figure*}
    \centering
    \includegraphics[width=\linewidth]{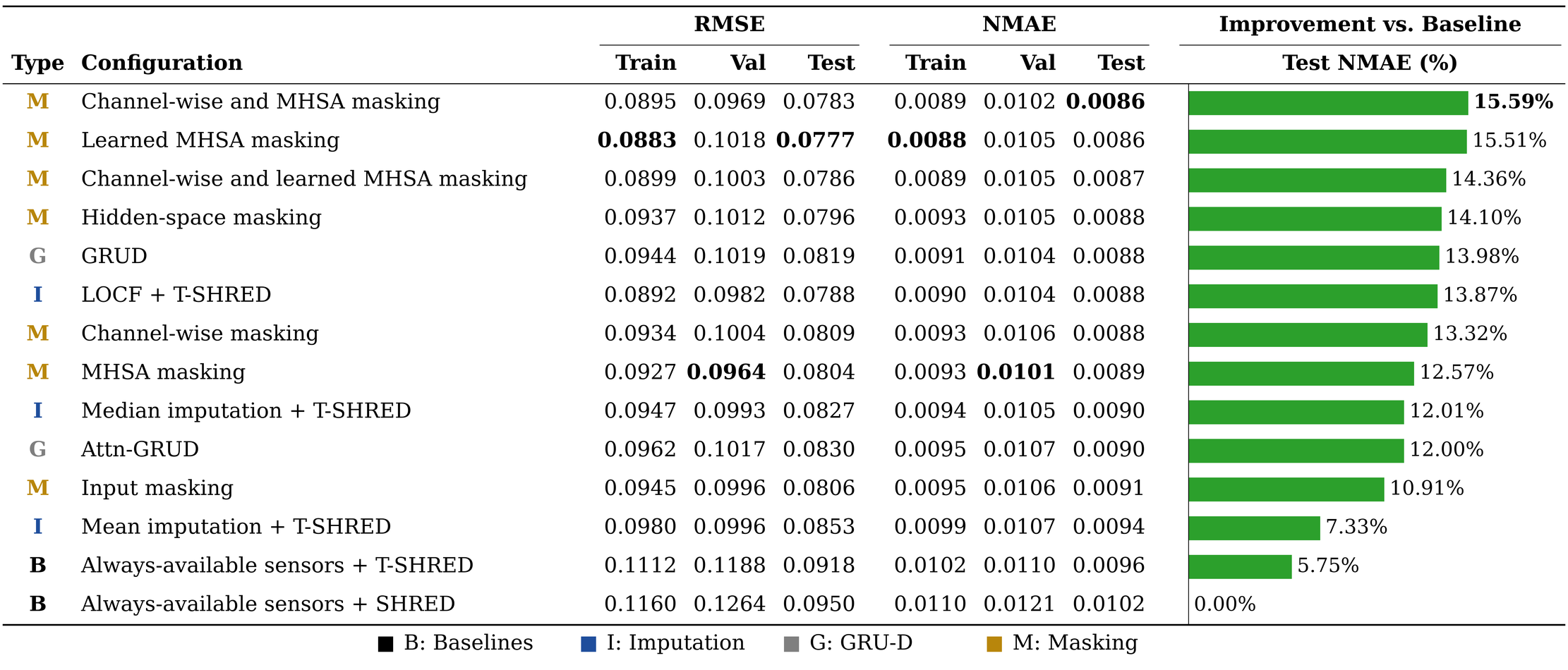}
    \caption{{\em Southern North Sea hydrodynamics}. Root mean squared errors and normalized mean absolute errors on training, validation and test data achieved by the different competing models.}
    \label{fig:hydro_table}
\end{figure*}
\begin{figure*}
    \centering
    \includegraphics[width=\linewidth]{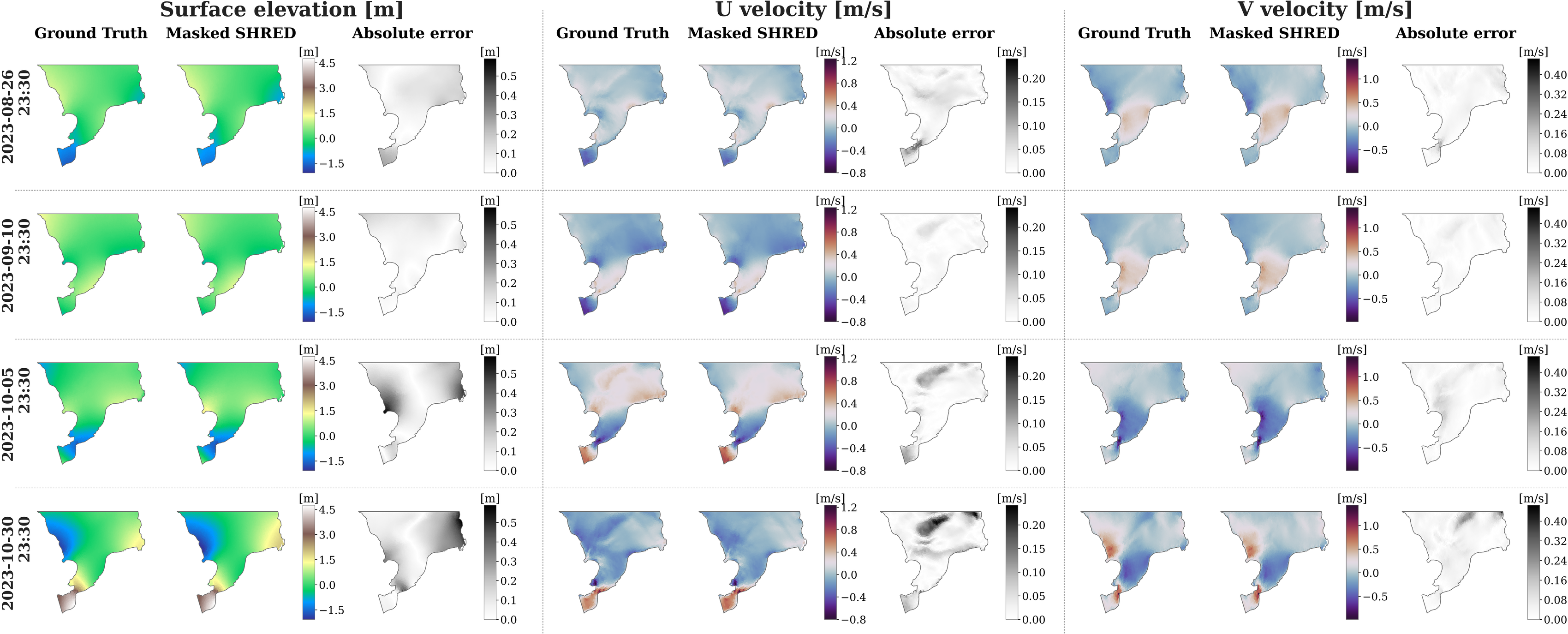}
    \caption{{\em Southern North Sea hydrodynamics}. Ground truth, reconstruction provided by masked SHRED with channel-wise and MHSA masking and reconstruction error on surface elevation (first column), horizontal velocity (second column) and vertical velocity (third column) at four different time instants in the test set.}
    \label{fig:hydro_results}
\end{figure*}
\begin{figure*}
    \centering
    \includegraphics[width=\linewidth]{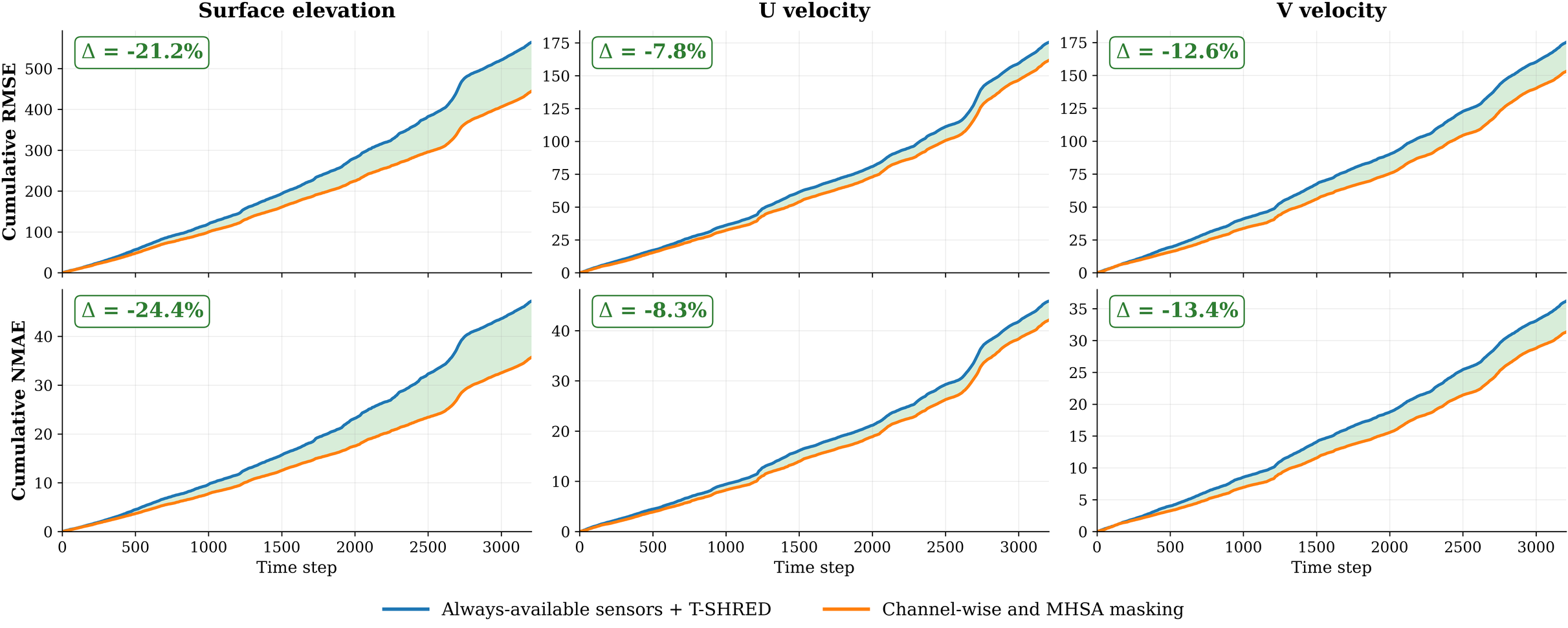}
    \caption{{\em Southern North Sea hydrodynamics}. Cumulative root mean squared error (first row) and normalized mean absolute error (second row) for the three output fields over the last two months in the test set, predicted by masked SHRED with channel-wise and MHSA masking and by the T-SHRED baseline trained exclusively on always-available sensors.}
    \label{fig:hydro_error_accumulation}
\end{figure*}
\begin{figure*}
    \centering
    \includegraphics[width=\linewidth]{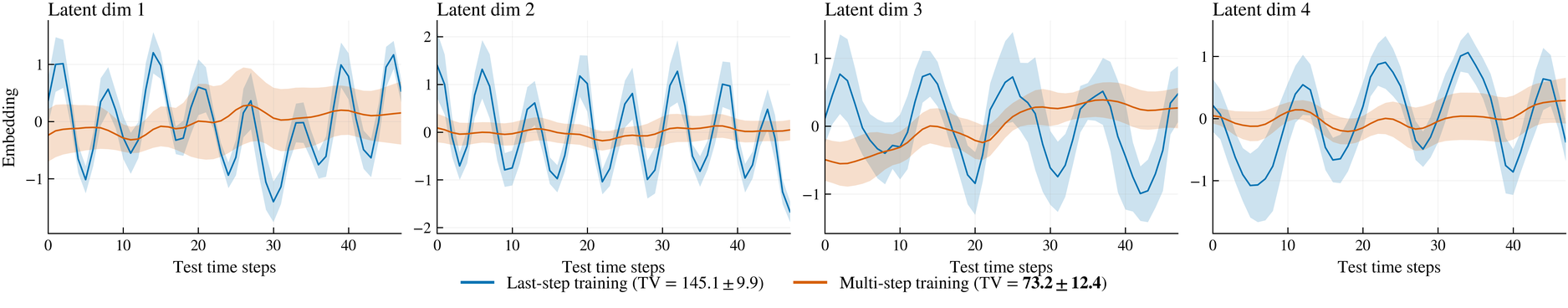}
    \caption{{\em Southern North Sea hydrodynamics}. Mean and one-standard-deviation intervals of the first four components of the latent variable $\tilde{\mathbf{z}}_k$ over time $k\in\Xi_{\text{test}}$ of masked SHRED with channel-wise and MHSA masking, evaluated using the different lagged sensor measurements in the test set and the last- and multi-step decoding training strategies.}
    \label{fig:hydro_embeddings}
\end{figure*}

This section delves into a first real-world application in the context of hydrodynamic modeling. Specifically, we aim at reconstructing the water level and velocity in the Southern North Sea relying solely on sensor readings provided by sparse water level stations within the region of interest. This challenging setting is characterized by spatial heterogeneity, complex coastal geometries, and strong temporal variability driven by tides, wind forcing, and boundary conditions, making it a suitable test case for assessing the robustness of the proposed strategies under heterogeneous sensor coverage. The high-dimensional surface elevation and current velocity fields ($N=25599$) from January 2022 to October 2023 are generated by numerically solving the two-dimensional shallow water equations with a $30$-minute resolution through the MIKE 21 Flow Model FM~\cite{HYDRO}, a two-dimensional, depth-averaged finite element model widely used in coastal and shelf-sea applications. The model employs observational water levels and current velocities data as boundary conditions, as well as wind velocities and air pressure data within the domain for calibration. Sensor measurements, instead, are provided by $16$ in situ measurement stations in the Southern North Sea~\cite{hydro_data, HYDRO}. The spatial coverage of water level stations is available in panel (a) of Figure~\ref{fig:hydro_availability}, while the corresponding data streams are available in panel (b) of the same figure. In particular, some stations exhibit intermittent gaps due to communication failures or acquisition issues, while others experience extended inactivity caused by hardware malfunctions or maintenance. Moreover, sensor readings are affected by missing values and uncertainties typical of operational ocean observing systems, including measurement noise, irregular sampling, communication delays, sensor outages, calibration errors, and local disturbances. It is worth noting that the sensor measurements do not exactly match the simulated field values at the corresponding station locations, although they exhibit a strong correlation~\cite{HYDRO}.

We here investigate the performance of the different competing methods in the ultra-sparse sensing regime where we challenge the models to reconstruct the water level and velocity in the Southern North Sea over time relying solely on $4$ in situ measurement stations, namely \texttt{ZeebruggeWielingendok}, \texttt{K141}, \texttt{NWA}, \texttt{Dover}, showing temporal coverage between $100\%$ and $48.6\%$, and coefficient of determination $R^2$ between sensor and pointwise simulation values between $0.934$ and $0.911$. After splitting the available data into training, validation and test with ratio $80:10:10$, so that the test set includes both months within and beyond the temporal range considered during training, we compress the surface elevation and velocity components through POD with, respectively, reduced dimensions equal to $12$, $40$ and $68$ POD modes, ensuring that the residual energy remains below $0.2\%$. Moreover, we consider sensor values lagged by $24$ time steps, which corresponds to a half-day delay.

In addition to GRU-D-based models, classical imputation methods and masking strategies in the context of T-SHRED, we benchmark the performance against SHRED with LSTM as recurrent temporal encoder and T-SHRED trained exclusively on the always-available sensor station, i.e. \texttt{ZeebruggeWielingendok}. To fairly compare the different fault-tolerant strategies, we consider the same training setup and architecture for all the methods. In particular, we take into account recurrent temporal encoders with $3$ layers and hidden size $32$, while Transformer-based encoders consider $3$ blocks, $4$ attention heads, embedding dimension $32$, and a MLP with a single hidden layer with $48$ neurons. The spatial decoder is instead modeled via a SDN with two hidden layers having $64$ and $96$ neurons, respectively. 

Figure~\ref{fig:hydro_table} displays the RMSE and the NMAE on training, validation and test data of the different baselines and fault-tolerant sensing strategies considered. Overall, the Transformer-based masking strategies proposed in Section~\ref{sec:masking:mshred} are able to provide the most accurate results on test data, i.e. when predicting the solution on time instants unseen during training, both within the training range and when extrapolating in time. The best-performing configuration, that is masked SHRED with channel-wise and MHSA masking, achieves a NMAE on test data equal to $0.86\%$, with a relative improvement of $15.59\%$ with respect to the SHRED baseline fitted on the always-available water level station only. Importantly, the limited margin of improvement in the real-world setting is consistent with the higher intrinsic difficulty of the task, the limited number of retained sensors, and the noise inherent in operational measurements, all of which compress the achievable performance range across competing methods. Figure~\ref{fig:hydro_results} displays, instead, the ground truth and the reconstructed fields at four different time instants in the test set provided by the best-performing fault-tolerant strategy when extrapolating in time with respect to the training data. Masked SHRED is able to provide accurate reconstructions of the different fields of interest when dealing with sensor values unseen during training. Moreover, by leveraging all the water level stations available, masked SHRED accumulates substantially less error in long-horizon prediction regimes, i.e. over the last two months in the test set, with respect to the T-SHRED baseline, as visible in Figure~\ref{fig:hydro_error_accumulation}.

The multi-step decoding training strategy proposed in Section~\ref{sec:masking:lagged} provides comparable results compared to the classical last-step decoding objective when applied to the different sensing strategies. Moreover, it yields smoother and temporally consistent latent trajectories $\tilde{\mathbf{z}}_k$ for $k \in \Xi_\text{test}$ compared to the ones produced by last-step training, as demonstrated by Figure~\ref{fig:hydro_embeddings} through the mean and standard deviation of the first four components of $\tilde{\mathbf{z}}_k$ computed across different lagged sensor values in the test set, with reference to masked SHRED with channel-wise and MHSA masking, thus potentially making latent forecasting more tractable and stable under distributional shift. To quantitatively support this finding, Figure~\ref{fig:hydro_embeddings} also reports the mean and the standard deviation of the total variation (TV) $\lVert \tilde{\mathbf{z}}_{k+1} - \tilde{\mathbf{z}}_k \rVert$ for $k \in \Xi_{\text{test}} \setminus \{m\}$. In particular, we observe a $49.6 \%$ reduction of the TV between the two training strategies, and a $25.1 \%$ reduction when considering the TV computed on the last-step embeddings $\mathbf{z}_k$ for $k \in \Xi_\text{test}$ only. 

\subsection{Øresund Strait hydrodynamics}
\label{sec:results:oresund}

\begin{figure*}
    \centering
    \includegraphics[width=\linewidth]{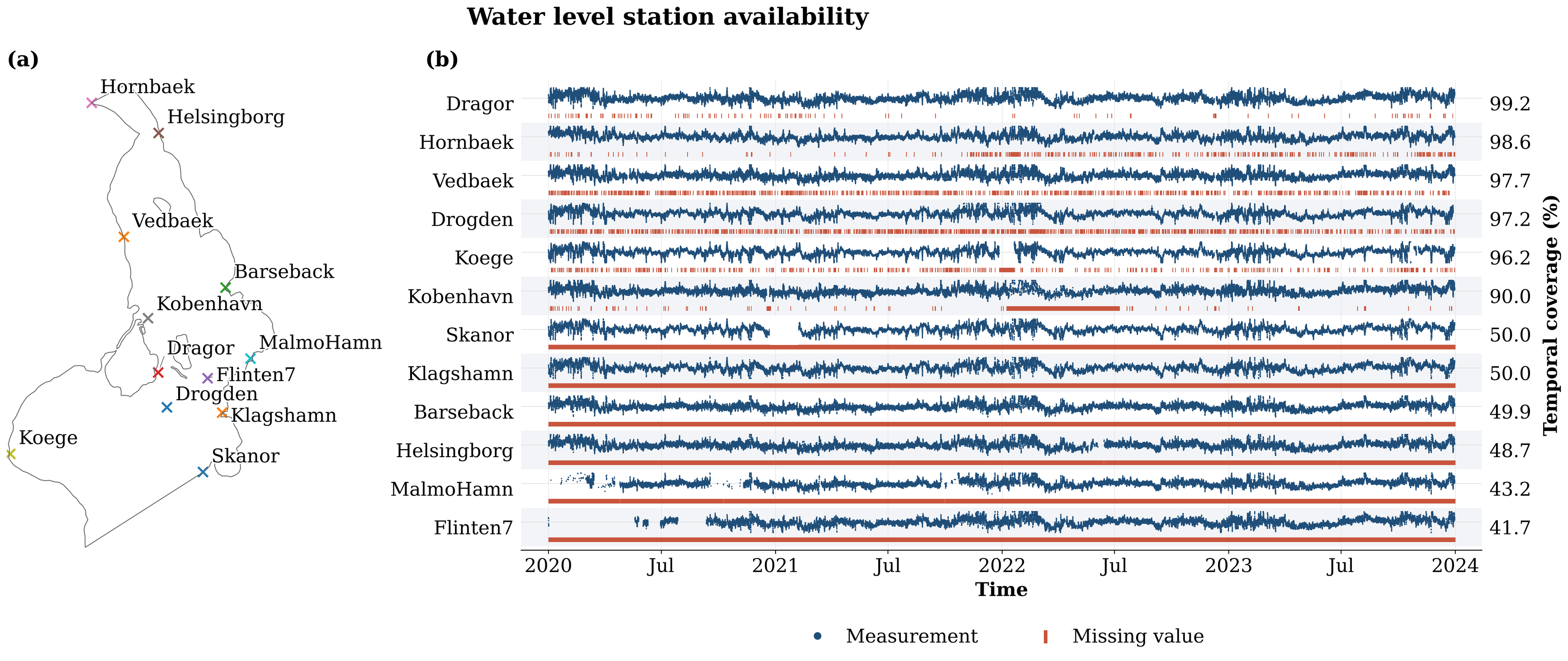}

    \caption{{\em Øresund Strait hydrodynamics}. {\bf (a)} spatial coverage of in situ water level stations in the Øresund Strait~\cite{ORESUND}. {\bf (b)} Sensor data availability over time of different in situ water level stations in the Øresund Strait~\cite{ORESUND}, covering the period from January 2020 to October 2023, showing availability rates ranging from $99.2\%$ to $41.7\%$ and different data availability patterns. The available sensor readings are depicted by blue dots, while missing values are represented by red bars.}
    \label{fig:oresund_availability}
\end{figure*}
\begin{figure*}
    \centering
    \includegraphics[width=\linewidth]{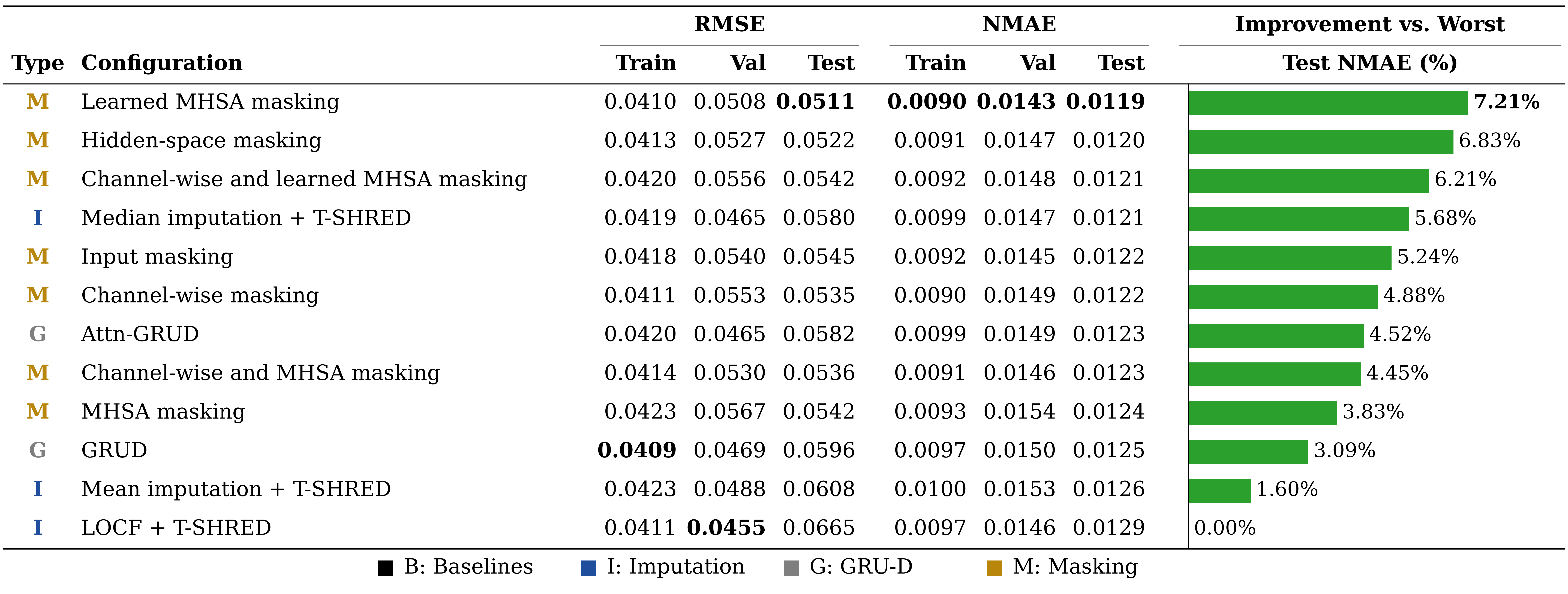}
    \caption{{\em Øresund Strait hydrodynamics}. Root mean squared errors and normalized mean absolute errors on training, validation and test data achieved by the different competing models.}
    \label{fig:oresund_table}
\end{figure*}

This section presents another real-world application where, similarly to the test case detailed in Section~\ref{sec:results:hydro}, we aim at reconstructing the water level and velocity in the Øresund Strait, an area of approximately $2000 \mathrm{km}^2$ between Denmark and Sweden, while relying solely on observational data provided by $12$ measurement stations~\cite{oresund_data, ORESUND}. The high-dimensional target fields ($N=9980$) from January 2020 to October 2023, that are the surface elevation and the current velocity solving the two-dimensional shallow water equations with a $30$-minute resolution, are provided by the MIKE 21 Flow Model FM~\cite{ORESUND}. The panel (a) of Figure~\ref{fig:oresund_availability} shows the spatial coverage of the available in situ measurements stations, while panel (b) of the same figure displays the data availability patterns of the different stations. For instance, \texttt{Koege} station does not provide measurements for about one month at the beginning of 2022, while the six stations with the lowest temporal coverage are characterized by a halved sampling rate with respect to those with higher temporal coverage. Notably, since no water level station provides a temporal coverage equal to $100\%$, it is not possible to consider naive SHRED baselines fitted on always-available sensors only. Similarly to the previous test case, the sensor values are not directly extracted from the simulated high-dimensional water level data, although they exhibit a strong correlation~\cite{ORESUND}.

After splitting the available data into training, validation and test with ratio $70:15:15$, we employ POD to compress the surface elevation and velocity components with reduced dimensions equal to, respectively, $15$, $100$ and $100$, so that the residual energy remains below $0.1\%$. Moreover, we consider a lag equal to $24$, which corresponds to a half-day delay. As far as the different competing models are concerned, we consider recurrent temporal encoders with $3$ hidden layers and hidden size equal to $32$, while Transformer-based encoders consider $3$ blocks, $4$ attention heads, embedding dimension $32$, and a MLP with a single hidden layer of dimension $48$. The spatial decoder is instead modeled via a SDN with two hidden layers having $64$ and $96$ neurons, where dropout with rate $0.1$ is employed to prevent overfitting.

\begin{figure*}
    \centering
    \includegraphics[width=\linewidth]{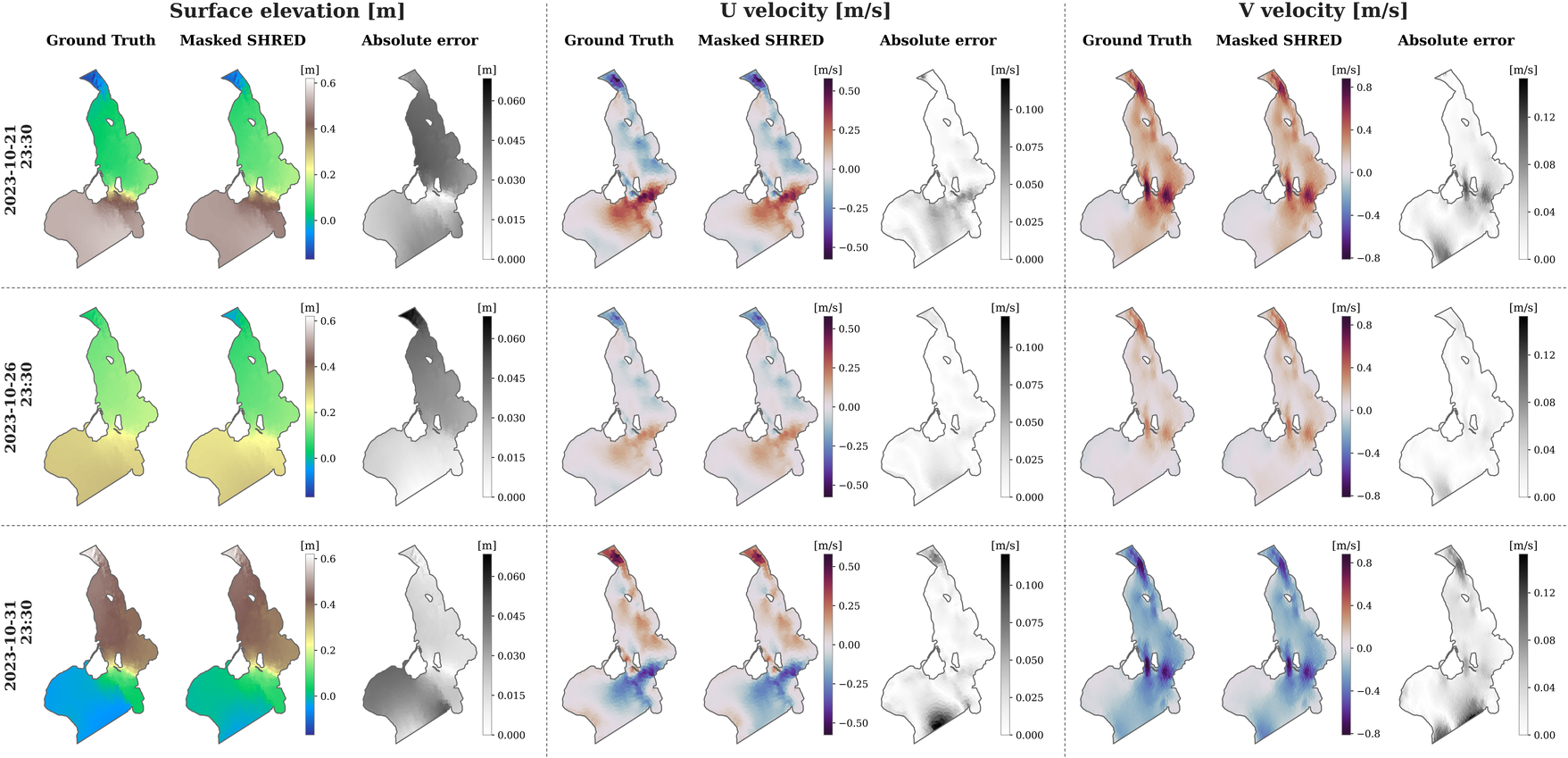}
    \caption{{\em Øresund Strait hydrodynamics}. Ground truth, reconstruction provided by masked SHRED with learned MHSA masking and reconstruction error on surface elevation (first column), horizontal velocity (second column) and vertical velocity (third column) at three different time instants in the test set.}
    \label{fig:oresund_results}
\end{figure*}

The error metrics, namely RMSE and NMAE, on training, validation and test set of the different competing methods are available in Figure~\ref{fig:oresund_table}. In particular, it is possible to assess the superiority of Transformer-based masking strategies with respect to imputation- and GRU-D-based techniques, although all strategies achieve comparable and acceptable performance. For example, the best-performing model, that is masked SHRED with learned MHSA masking, provides a NMAE on test data equal to $1.19 \%$. To qualitatively assess the goodness of the predictions provided by the best-performing model, Figure~\ref{fig:oresund_results} displays the ground truth and the corresponding reconstruction at three different time instants in the test set when extrapolating with respect to the training time range.

\subsection{Sea surface temperature}
\label{sec:results:sst}

The last real-world application deals with sea surface temperature data, where the high-resolution snapshots ($N=64800$) from January 2000 to December 2025 are obtained by combining daily satellite observations on a regular grid with in situ measurements from different ocean-observing platforms, such as ships and buoys, to fill possible gaps in observational coverage and correct systematic biases~\cite{sst_data}. The goal of this test case is to predict the spatio-temporal high-resolution SST field over the entire globe relying solely on a limited amount of observational data provided by $20$ moored buoys monitoring SST within the domain and over the same time horizon. Figure~\ref{fig:availability}, panel (a), shows the spatial coverage of the considered moored buoys, while the panel (b) of the same figure displays the patterns of missing values of the SST buoy data.

\begin{figure*}
    \centering
    \includegraphics[width=\linewidth]{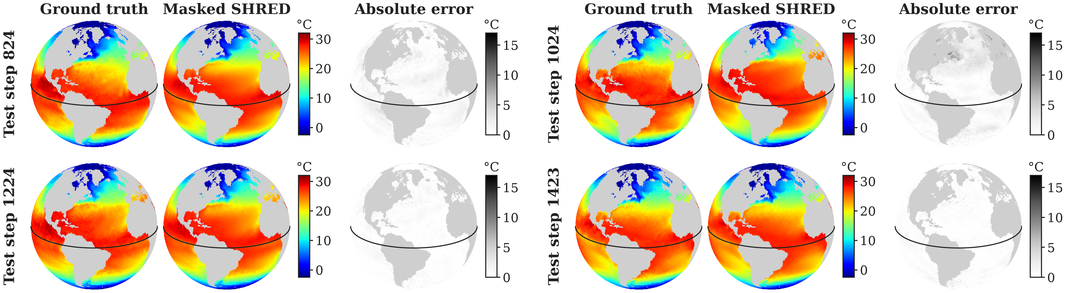}
    \caption{{\em Sea surface temperature}. Ground truth, reconstruction provided by masked SHRED with channel-wise and learned MHSA masking and reconstruction error on sea surface temperature at four different time instants in the test set.}
    \label{fig:sst_results}
\end{figure*}
\begin{figure*}
    \centering
    \includegraphics[width=\linewidth]{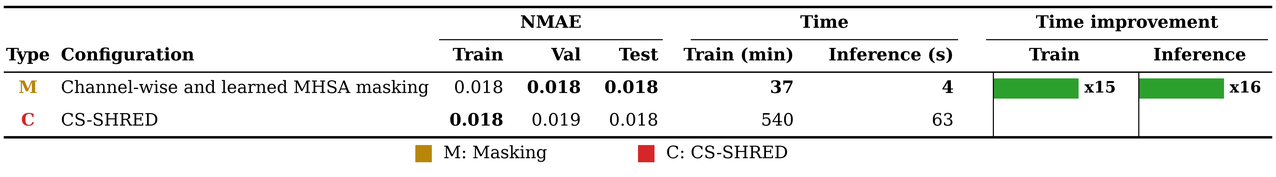}
    \caption{{\em Sea surface temperature}. Normalized mean absolute errors on training, validation and test data (first column) and training and inference times (second column) of the different competing models.}
    \label{fig:sst_table}
\end{figure*}

After splitting the available data into training, validation and test with ratio $70:15:15$, we reduced the dimensionality of the high-resolution SST snapshots through POD by considering a reduced dimension equal to $500$. Moreover, we employ a lag equal to $40$, which corresponds to $40$ days of historical data, and we consider $400$ epochs for training. Regarding the model architectures, the recurrent temporal encoders are made of one single hidden layer of size $32$, while Transformer-based encoders employ $1$ block, $4$ attention heads, embedding dimension $32$, and a MLP with a single hidden layer of dimension $48$. The spatial decoder is instead modeled via a SDN with a single hidden layer of size $256$, where dropout with rate $0.1$ is employed to prevent overfitting.

Figure~\ref{fig:sst_results} shows the ground truth and the reconstructed SST fields provided by masked SHRED with channel-wise and learned MHSA masking at four different time instants in the test set when extrapolating over time with respect to the training data. Figure~\ref{fig:sst_table}, instead, compares the proposed masked SHRED strategy with CS-SHRED~\cite{SHRED_L1}: while both solutions are able to cope with missing sensor measurements and achieve acceptable results on the test set, masked SHRED is far more efficient both in training ($15\times$ of speed up) and during evaluation ($16\times$ of speed up), since it does not require computationally expensive convex minimization problems to recover missing values and incurs no additional computational overhead with respect to standard SHRED implementation.

While all competing methods are evaluated using the same hyperparameter values throughout the test cases discussed in Section~\ref{sec:results}, it is worth noting that additional experiments involving hyperparameter tuning yield similar results and support the same overall conclusions. In particular, masked SHRED systematically outperforms imputation- and GRU-D-based methods, with only minor variations in the ranking of the different masking strategies. Indeed, the proposed Transformer-based masking strategies exhibit limited sensitivity to hyperparameter variations, with their effectiveness largely preserved across different hyperparameter values.

\section{CONCLUSIONS}
\label{sec:conclusion}
This work addresses one of the main challenges in the context of sparse sensing, namely the ability to handle incomplete, heterogeneous, and noisy sensor measurements that characterize operational sensing environments. Indeed, state-of-the-art sparse sensing techniques such as, e.g., shallow recurrent decoders (SHRED), typically consider continuously active sensors with the same sampling rates, that are often directly extracted as pointwise values of the corresponding full-order state. To move toward practical settings, we first introduce a hierarchy of increasingly expressive strategies, ranging from standard imputation baselines to recurrent architectures specifically designed for irregular time-series forecasting. We then propose four Transformer-based masking strategies in order to make the SHRED temporal encoder explicitly aware of sensor availability patterns. Specifically, these strategies inject the binary data-availability mask into the Transformer-based encoder at different levels of the architecture, including input concatenation (channel-wise masking), hidden-state gating (hidden-space masking), and additive attention biases within the multi-head self-attention mechanism, either fixed or learned (MHSA and learned MHSA masking). Finally, we introduce a multi-step decoding training strategy that enforces structured latent representations throughout the full observation window.

The different state-of-the-art imputation methods and the proposed masked shallow recurrent decoders have been employed on four different test cases to assess their performance. In particular, we consider a synthetic benchmark on shallow water equation solution data, where sensors are directly extracted from the corresponding full-order state and sensor malfunctions are simulated for the sake of the analysis. Moreover, we take into account three real-world operational datasets in the context of hydrodynamic and ocean modeling, thus assessing the performance of the different proposed strategies on realistic, intermittent and asynchronous sensor values. Across all the test cases considered, the fault-tolerant techniques considered improve the naive SHRED baseline relying solely on the always active sensors. Moreover, masked shallow recurrent decoders consistently provide accurate state reconstructions on new time steps never seen during training, in both interpolation and extrapolation regimes, yielding lower errors than both imputation baselines and GRU-D-based approaches, with no additional inference-time computational cost. Moreover, the multi-step decoding training scheme improves upon the last-step training baseline in terms of predictive accuracy, and results in substantially smoother latent trajectories, thus potentially making latent forecasting more tractable and stable under distributional shift. Note that, in general, the different masking strategies proposed may be combined and there is no universally superior masking solution. However, the hybrid combinations are often ranked among the top-performing configurations across the different test cases, suggesting that the injection of the binary data-availability mask at multiple levels of the architecture may be beneficial.

The fault-tolerant shallow recurrent decoders introduced in this work may be extended in multiple directions in future works. For instance, one may consider an ensemble of masked SHRED models to provide robust estimates under noisy and intermittent sensor measurements, similarly to \cite{riva2024robuststateestimationpartial}. Moreover, it is possible to consider parametric problems where fault-tolerant SHRED must generalize across varying physical and geometrical parameters, in the direction of SHRED-ROM~\cite{SHRED-ROM}. In this setting, the fault-tolerant SHRED architecture could additionally support parameter estimation and uncertainty propagation from incomplete sensor measurements to reconstructed fields. Finally, the smoother embeddings produced by the multi-step decoding training strategy may be exploited to accurately forecast the system evolution at the latent level, thus enabling masked SHRED to effectively predict full-order states in advance even when the sensors are not yet available.

\vfill
\section*{CODE AVAILABILITY} 
The code can be found in our repository \\
\url{https://github.com/matteopasqual02/Masked-SHRED}.

\section*{ACKNOWLEDGEMENTS} 
MT thanks Jesper Sandvig Mariegaard and Freja Høgholm Petersen for the fruitful discussions and valuable insights on hydrodynamic modeling and real-world sensing. AM acknowledges the Project “Reduced Order Modeling and Deep Learning for the real-time approximation of PDEs (DREAM)” (Starting Grant No. FIS00003154), funded by the Italian Science Fund (FIS) - Ministero dell'Università e della Ricerca and the project FAIR (Future Artificial Intelligence Research), funded by the NextGenerationEU program within the PNRR-PE-AI scheme (M4C2, Investment 1.3, Line on Artificial Intelligence).

\vfill
\vspace{-0.5cm}
\onecolumngrid
\vspace{5mm}
\noindent\centering\rule{0.5\textwidth}{1pt}
\vspace{5mm}

\twocolumngrid
\renewcommand{\bibsection}{}
\bibliographystyle{abbrv}
\bibliography{references}

\end{document}